\documentclass[10pt,twocolumn,letterpaper]{article}

\usepackage{cvpr}              

\makeatletter
\@namedef{ver@everyshi.sty}{}
\makeatother

\usepackage[dvipsnames,svgnames,x11names]{xcolor}
\usepackage{tikz}
\usetikzlibrary{arrows.meta,shapes,calc,matrix,fit,positioning,backgrounds,decorations.markings}
\usepackage{pgfplots}
\usepackage{pgfplotstable}
\pgfplotsset{compat=1.9}
\usepackage{xstring}

\usepgfplotslibrary{external}

\IfBeginWith*{\jobname}{fig/extern/}{\finalcopy}{}

\tikzstyle{every picture}+=[
	remember picture,
	every text node part/.style={align=center},
	every matrix/.append style={ampersand replacement=\&},
]
\tikzstyle{tight} = [inner sep=0pt,outer sep=0pt]
\tikzstyle{node}  = [draw,circle,tight,minimum size=12pt,anchor=center]
\tikzstyle{op}    = [draw,circle,tight]
\tikzstyle{dot}   = [fill,draw,circle,inner sep=1pt,outer sep=0]
\tikzstyle{pt}    = [fill,draw,circle,inner sep=1.5pt,outer sep=.2pt]
\tikzstyle{box}   = [draw,rectangle,inner sep=3pt]
\tikzstyle{high}  = [black!60]
\tikzstyle{group} = [high,box,opacity=.5]
\tikzstyle{dim1}  = [fill opacity=.3,text opacity=1]
\tikzstyle{dim2}  = [fill opacity=.5,text opacity=1]
\tikzstyle{dim3}  = [fill opacity=.7,text opacity=1]
\tikzstyle{rectc} = [tight,transform shape]
\tikzstyle{rect}  = [rectc,anchor=south west]

\tikzset{every mark/.append style={solid}}
\pgfplotsset{
	grid=both, width=\columnwidth, try min ticks=5,
	every axis/.append style={font=\small},
	every axis plot/.append style={thick,mark=none,mark size=1.8,tension=0.18},
	legend cell align=left, legend style={fill opacity=0.8},
	xticklabel={\pgfmathprintnumber[assume math mode=true]{\tick}},
	yticklabel={\pgfmathprintnumber[assume math mode=true]{\tick}},
	nodes near coords math/.style={
		nodes near coords={\pgfmathprintnumber[assume math mode=true]{\pgfplotspointmeta}},
	},
}

\pgfplotsset{
	dash/.style={mark=o,dashed,opacity=0.6},
	dott/.style={mark=o,dotted,opacity=0.6},
	nolim/.style={enlargelimits=false},
	plain/.style={every axis plot/.append style={},nolim,grid=none},
}

\pgfdeclarelayer{bg4}
\pgfdeclarelayer{bg3}
\pgfdeclarelayer{bg2}
\pgfdeclarelayer{bg1}
\pgfdeclarelayer{fg1}
\pgfdeclarelayer{fg2}
\pgfdeclarelayer{fg3}
\pgfdeclarelayer{fg4}
\pgfsetlayers{bg4,bg3,bg2,bg1,main,fg1,fg2,fg3,fg4}

\pgfkeys{/tikz/.cd, aspect/.store in=\aspect, aspect=1}
\pgfkeys{/tikz/.cd, depth/.store in=\depth, depth=.5}
\pgfkeys{/tikz/.cd, stepx/.store in=\step, stepx=1}

\tikzstyle{geom} = [line join=bevel,aspect=1,depth=.5,z={(\depth*\aspect,\depth)}]
\tikzstyle{wire} = [geom,draw,thick]

\def\cx[#1,#2,#3]{#1}
\def\cy[#1,#2,#3]{#2}
\def\cz[#1,#2,#3]{#3}
\def\ex[#1,#2,#3]{#1,0,0}
\def\ey[#1,#2,#3]{0,#2,0}
\def\ez[#1,#2,#3]{0,0,#3}

\makeatletter
\renewcommand\paragraph{\@startsection{paragraph}{4}{\z@}{.6ex}{-1em}{\normalfont\normalsize\bfseries}}
\makeatother

\usepackage{times}
\usepackage{epsfig}
\usepackage{graphicx}
\usepackage{amsmath}
\usepackage{amssymb}
\usepackage{booktabs}
\usepackage{multirow}
\usepackage{comment}
\usepackage{color}
\usepackage[accsupp]{axessibility}
\usepackage{makecell}

\usepackage{mathtools}

\usepackage{enumitem}
\usepackage{xspace}
\usepackage{breakcites}
\usepackage{contour}
\usepackage{pifont}

\usepackage[utf8]{inputenc}
\usepackage{kotex}
\usepackage{kotex-logo}

\usepackage{algorithm}
\usepackage{algpseudocode}

\usepackage{mycommands}

\usepackage[pagebackref=true,breaklinks=true,colorlinks,bookmarks=false]{hyperref}

\usepackage[capitalize]{cleveref}
\crefname{section}{Sec.}{Secs.}
\Crefname{section}{Section}{Sections}
\Crefname{table}{Table}{Tables}
\crefname{table}{Tab.}{Tabs.}

\def\cvprPaperID{9111} 
\def\confName{CVPR}
\def\confYear{2023}

\begin{document}

\title{On Train-Test Class Overlap and Detection for Image Retrieval}

\author{
Chull Hwan Song$^1$ \ \ \ \ Jooyoung Yoon$^1$ \ \ \ \ Taebaek Hwang$^1$\ \ \ \ Shunghyun Choi$^1$ \\ Yeong Hyeon Gu$^{2}$\thanks{Corresponding author}\ \ \ \ Yannis Avrithis$^{3}$\\
\normalsize $^1$Dealicious Inc.\quad
\normalsize $^2$Sejong University \quad
\normalsize $^3$Institute of Advanced Research on Artificial Intelligence (IARAI)\\
}

\maketitle

\thispagestyle{empty}

\newcommand{\head}[1]{{\smallskip\noindent\textbf{#1}}}
\newcommand{\alert}[1]{{\color{red}{#1}}}
\newcommand{\sm}{\scriptsize}
\newcommand{\eq}[1]{(\ref{eq:#1})}

\newcommand{\Th}[1]{\textsc{#1}}
\newcommand{\mr}[2]{\multirow{#1}{*}{#2}}
\newcommand{\mc}[2]{\multicolumn{#1}{c}{#2}}
\newcommand{\mca}[3]{\multicolumn{#1}{#2}{#3}}
\newcommand{\tb}[1]{\textbf{#1}}
\newcommand{\ch}{\checkmark}

\newcommand{\red}[1]{{\color{red}{#1}}}
\newcommand{\blue}[1]{{\color{blue}{#1}}}
\newcommand{\green}[1]{\color{green}{#1}}
\newcommand{\gray}[1]{{\color{gray}{#1}}}
\newcommand{\orange}[1]{{\color{orange}{#1}}}
\definecolor{lightgray2}{rgb}{0.9, 0.9, 0.9}

\newcommand{\citeme}[1]{\red{[XX]}}
\newcommand{\refme}[1]{\red{(XX)}}

\newcommand{\fig}[2][1]{\includegraphics[width=#1\linewidth]{fig/#2}}
\newcommand{\figh}[2][1]{\includegraphics[height=#1\linewidth]{fig/#2}}
\newcommand{\figwh}[3]{\includegraphics[width=#1\linewidth,height=#2\linewidth]{fig/#3}}


\newcommand{\tran}{^\top}
\newcommand{\mtran}{^{-\top}}
\newcommand{\zcol}{\mathbf{0}}
\newcommand{\zrow}{\zcol\tran}

\newcommand{\ind}{\mathbbm{1}}
\newcommand{\expect}{\mathbb{E}}
\newcommand{\nat}{\mathbb{N}}
\newcommand{\zahl}{\mathbb{Z}}
\newcommand{\real}{\mathbb{R}}
\newcommand{\proj}{\mathbb{P}}
\newcommand{\prob}{\mathbf{Pr}}
\newcommand{\normal}{\mathcal{N}}

\newcommand{\mif}{\textrm{if}\ }
\newcommand{\other}{\textrm{otherwise}}
\newcommand{\minimize}{\textrm{minimize}\ }
\newcommand{\maximize}{\textrm{maximize}\ }
\newcommand{\st}{\textrm{subject\ to}\ }

\newcommand{\id}{\operatorname{id}}
\newcommand{\const}{\operatorname{const}}
\newcommand{\sgn}{\operatorname{sgn}}
\newcommand{\var}{\operatorname{Var}}
\newcommand{\mean}{\operatorname{mean}}
\newcommand{\trace}{\operatorname{tr}}
\newcommand{\diag}{\operatorname{diag}}
\newcommand{\vect}{\operatorname{vec}}
\newcommand{\cov}{\operatorname{cov}}
\newcommand{\sign}{\operatorname{sign}}
\newcommand{\prj}{\operatorname{proj}}

\newcommand{\sigmoid}{\operatorname{sigmoid}}
\newcommand{\softmax}{\operatorname{softmax}}
\newcommand{\clip}{\operatorname{clip}}

\newcommand{\defn}{\mathrel{:=}}
\newcommand{\peq}{\mathrel{+\!=}}
\newcommand{\meq}{\mathrel{-\!=}}

\newcommand{\floor}[1]{\left\lfloor{#1}\right\rfloor}
\newcommand{\ceil}[1]{\left\lceil{#1}\right\rceil}
\newcommand{\inner}[1]{\left\langle{#1}\right\rangle}
\newcommand{\norm}[1]{\left\|{#1}\right\|}
\newcommand{\abs}[1]{\left|{#1}\right|}
\newcommand{\frob}[1]{\norm{#1}_F}
\newcommand{\card}[1]{\left|{#1}\right|\xspace}
\newcommand{\diff}{\mathrm{d}}
\newcommand{\der}[3][]{\frac{d^{#1}#2}{d#3^{#1}}}
\newcommand{\pder}[3][]{\frac{\partial^{#1}{#2}}{\partial{#3^{#1}}}}
\newcommand{\ipder}[3][]{\partial^{#1}{#2}/\partial{#3^{#1}}}
\newcommand{\dder}[3]{\frac{\partial^2{#1}}{\partial{#2}\partial{#3}}}

\newcommand{\wb}[1]{\overline{#1}}
\newcommand{\wt}[1]{\widetilde{#1}}

\def\xssp{\hspace{1pt}}
\def\ssp{\hspace{3pt}}
\def\msp{\hspace{5pt}}
\def\lsp{\hspace{12pt}}

\newcommand{\cA}{\mathcal{A}}
\newcommand{\cB}{\mathcal{B}}
\newcommand{\cC}{\mathcal{C}}
\newcommand{\cD}{\mathcal{D}}
\newcommand{\cE}{\mathcal{E}}
\newcommand{\cF}{\mathcal{F}}
\newcommand{\cG}{\mathcal{G}}
\newcommand{\cH}{\mathcal{H}}
\newcommand{\cI}{\mathcal{I}}
\newcommand{\cJ}{\mathcal{J}}
\newcommand{\cK}{\mathcal{K}}
\newcommand{\cL}{\mathcal{L}}
\newcommand{\cM}{\mathcal{M}}
\newcommand{\cN}{\mathcal{N}}
\newcommand{\cO}{\mathcal{O}}
\newcommand{\cP}{\mathcal{P}}
\newcommand{\cQ}{\mathcal{Q}}
\newcommand{\cR}{\mathcal{R}}
\newcommand{\cS}{\mathcal{S}}
\newcommand{\cT}{\mathcal{T}}
\newcommand{\cU}{\mathcal{U}}
\newcommand{\cV}{\mathcal{V}}
\newcommand{\cW}{\mathcal{W}}
\newcommand{\cX}{\mathcal{X}}
\newcommand{\cY}{\mathcal{Y}}
\newcommand{\cZ}{\mathcal{Z}}

\newcommand{\vA}{\mathbf{A}}
\newcommand{\vB}{\mathbf{B}}
\newcommand{\vC}{\mathbf{C}}
\newcommand{\vD}{\mathbf{D}}
\newcommand{\vE}{\mathbf{E}}
\newcommand{\vF}{\mathbf{F}}
\newcommand{\vG}{\mathbf{G}}
\newcommand{\vH}{\mathbf{H}}
\newcommand{\vI}{\mathbf{I}}
\newcommand{\vJ}{\mathbf{J}}
\newcommand{\vK}{\mathbf{K}}
\newcommand{\vL}{\mathbf{L}}
\newcommand{\vM}{\mathbf{M}}
\newcommand{\vN}{\mathbf{N}}
\newcommand{\vO}{\mathbf{O}}
\newcommand{\vP}{\mathbf{P}}
\newcommand{\vQ}{\mathbf{Q}}
\newcommand{\vR}{\mathbf{R}}
\newcommand{\vS}{\mathbf{S}}
\newcommand{\vT}{\mathbf{T}}
\newcommand{\vU}{\mathbf{U}}
\newcommand{\vV}{\mathbf{V}}
\newcommand{\vW}{\mathbf{W}}
\newcommand{\vX}{\mathbf{X}}
\newcommand{\vY}{\mathbf{Y}}
\newcommand{\vZ}{\mathbf{Z}}

\newcommand{\va}{\mathbf{a}}
\newcommand{\vb}{\mathbf{b}}
\newcommand{\vc}{\mathbf{c}}
\newcommand{\vd}{\mathbf{d}}
\newcommand{\ve}{\mathbf{e}}
\newcommand{\vf}{\mathbf{f}}
\newcommand{\vg}{\mathbf{g}}
\newcommand{\vh}{\mathbf{h}}
\newcommand{\vi}{\mathbf{i}}
\newcommand{\vj}{\mathbf{j}}
\newcommand{\vk}{\mathbf{k}}
\newcommand{\vl}{\mathbf{l}}
\newcommand{\vm}{\mathbf{m}}
\newcommand{\vn}{\mathbf{n}}
\newcommand{\vo}{\mathbf{o}}
\newcommand{\vp}{\mathbf{p}}
\newcommand{\vq}{\mathbf{q}}
\newcommand{\vr}{\mathbf{r}}
\newcommand{\Vs}{\mathbf{s}}
\newcommand{\vt}{\mathbf{t}}
\newcommand{\vu}{\mathbf{u}}
\newcommand{\vv}{\mathbf{v}}
\newcommand{\vw}{\mathbf{w}}
\newcommand{\vx}{\mathbf{x}}
\newcommand{\vy}{\mathbf{y}}
\newcommand{\vz}{\mathbf{z}}

\newcommand{\vone}{\mathbf{1}}
\newcommand{\vzero}{\mathbf{0}}

\newcommand{\valpha}{{\boldsymbol{\alpha}}}
\newcommand{\vbeta}{{\boldsymbol{\beta}}}
\newcommand{\vgamma}{{\boldsymbol{\gamma}}}
\newcommand{\vdelta}{{\boldsymbol{\delta}}}
\newcommand{\vepsilon}{{\boldsymbol{\epsilon}}}
\newcommand{\vzeta}{{\boldsymbol{\zeta}}}
\newcommand{\veta}{{\boldsymbol{\eta}}}
\newcommand{\vtheta}{{\boldsymbol{\theta}}}
\newcommand{\viota}{{\boldsymbol{\iota}}}
\newcommand{\vkappa}{{\boldsymbol{\kappa}}}
\newcommand{\vlambda}{{\boldsymbol{\lambda}}}
\newcommand{\vmu}{{\boldsymbol{\mu}}}
\newcommand{\vnu}{{\boldsymbol{\nu}}}
\newcommand{\vxi}{{\boldsymbol{\xi}}}
\newcommand{\vomikron}{{\boldsymbol{\omikron}}}
\newcommand{\vpi}{{\boldsymbol{\pi}}}
\newcommand{\vrho}{{\boldsymbol{\rho}}}
\newcommand{\vsigma}{{\boldsymbol{\sigma}}}
\newcommand{\vtau}{{\boldsymbol{\tau}}}
\newcommand{\vupsilon}{{\boldsymbol{\upsilon}}}
\newcommand{\vphi}{{\boldsymbol{\phi}}}
\newcommand{\vchi}{{\boldsymbol{\chi}}}
\newcommand{\vpsi}{{\boldsymbol{\psi}}}
\newcommand{\vomega}{{\boldsymbol{\omega}}}

\newcommand{\rLambda}{\mathrm{\Lambda}}
\newcommand{\rSigma}{\mathrm{\Sigma}}

\newcommand{\vLambda}{\bm{\rLambda}}
\newcommand{\vSigma}{\bm{\rSigma}}

\makeatletter
\newcommand*\bdot{\mathpalette\bdot@{.7}}
\newcommand*\bdot@[2]{\mathbin{\vcenter{\hbox{\scalebox{#2}{$\m@th#1\bullet$}}}}}
\makeatother

\makeatletter
\DeclareRobustCommand\onedot{\futurelet\@let@token\@onedot}
\def\@onedot{\ifx\@let@token.\else.\null\fi\xspace}

\def\eg{\emph{e.g}\onedot} \def\Eg{\emph{E.g}\onedot}
\def\ie{\emph{i.e}\onedot} \def\Ie{\emph{I.e}\onedot}
\def\cf{\emph{cf}\onedot} \def\Cf{\emph{Cf}\onedot}
\def\etc{\emph{etc}\onedot} \def\vs{\emph{vs}\onedot}
\def\wrt{w.r.t\onedot} \def\dof{d.o.f\onedot} \def\aka{a.k.a\onedot}
\def\etal{\emph{et al}\onedot}
\makeatother




\newcommand{\cls}{{\texttt{[CLS]}}\xspace}

\newcommand{\relu}{\operatorname{relu}}
\newcommand{\conv}{\operatorname{conv}}
\newcommand{\aconv}{\operatorname{aconv}}

\newcommand{\fc}{\textsc{fc}}
\newcommand{\gap}{\textsc{gap}}
\newcommand{\bn}{\textsc{bn}}
\newcommand{\dropout}{\textsc{dropout}}

\newcommand{\elm}{\textsc{elm}}
\newcommand{\irb}{\textsc{irb}}
\newcommand{\wav}{\textsc{wb}}
\newcommand{\aspp}{\textsc{aspp}}
\newcommand{\fuse}{\textsc{fuse}}

\newcommand{\ours}{CiDeR\xspace}
\newcommand{\oursf}{CiDeR-FT\xspace}


\def\vlad{VLAD\xspace}
\def\smk{SMK$^{\star}$\xspace}
\def\asmk{ASMK$^{\star}$\xspace}
\def\sp{SP\xspace}
\def\qe{QE\xspace}
\def\hqe{HQE\xspace}
\def\dfs{DFS\xspace}
\def\off{O}
\def\hesaff{HesAff\xspace}
\def\rsift{rSIFT\xspace}
\def\delf{DELF\xspace}

\def\oxf5k{Ox5k\xspace}
\def\paris6k{Par6k\xspace}
\def\roxf{$\cR$Oxford\xspace}
\def\rox{$\cR$Oxf\xspace}
\def\ro{$\cR$O\xspace}
\def\rpar{$\cR$Paris\xspace}
\def\rpa{$\cR$Par\xspace}
\def\rp{$\cR$P\xspace}
\def\r1m{$\cR$1M\xspace}
\def\rs{$\cR$100k\xspace}


\newcommand{\cw}{0.8cm}
\newcommand{\yes}{\ch}
\newcommand{\no}{}


\newcommand{\gain}[1]{{\color{green!60!black}#1}}

\definecolor{redd}{rgb}{0.3,0.2, 0.7}

\newcommand{\ok}[1]{\color{redd}{#1}}






%
%
%
%
%
%


 \newcommand{\iavr}[1]{\blue{[Yannis: #1]}}
 \newcommand{\CH}[1]{\orange{[CheolHwan: #1]}}

\begin{abstract}
	How important is it for training and evaluation sets to not have class overlap in image retrieval? We revisit Google Landmarks v2 clean~\cite{Weyand01}, the most popular training set, by identifying and removing class overlap with Revisited Oxford and Paris~\cite{RITAC18}, the most popular evaluation set. By comparing the original and the new $\cR$GLDv2-clean on a benchmark of reproduced state-of-the-art methods, our findings are striking. Not only is there a dramatic drop in performance, but it is inconsistent across methods, changing the ranking.

	What does it take to focus on objects or interest and ignore background clutter when indexing?
	Do we need to train an object detector and the representation separately? Do we need location supervision? We introduce Single-stage Detect-to-Retrieve (\ours), an end-to-end, single-stage pipeline to detect objects of interest and extract a global image representation. We outperform previous state-of-the-art on both existing training sets and the new $\cR$GLDv2-clean. Our dataset is available at \url{https://github.com/dealicious-inc/RGLDv2-clean}.
\end{abstract}

\vspace{-10pt}
\section{Introduction}
\label{sec:intro}

\begin{figure}[t]
\vspace{-14pt}
\centering
\scriptsize
\begin{tabular}{cc}
\figh[.67]{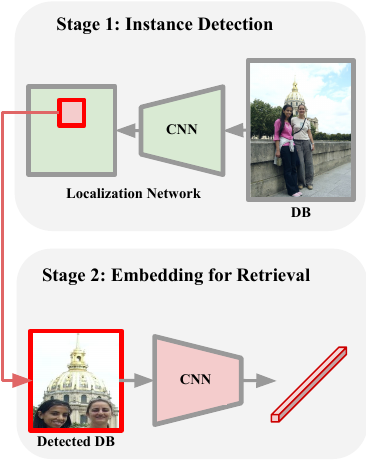} & 
\figh[.67]{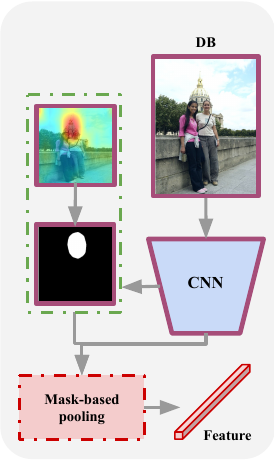} \\ 
(a) Two-stage &
(b) One-stage \\
\end{tabular}
\vspace{-7pt} 
\caption{It is beneficial for image retrieval to detect objects of interest in database images and only represent those. (a) \emph{Two-stage} pipeline. Previous works involve two-stage embedding extraction at indexing, or a two-stage training process, and they may use location supervision or not. (b) \emph{One-stage} pipeline. We use a single-stage embedding extraction at training and indexing; training is end-to-end and uses no location supervision.}
\label{fig:fig1}
\end{figure}

Instance-level image retrieval is a significant computer vision problem, attracting substantial investigation before and after deep learning. High-quality datasets are crucial for advancing research. Image retrieval has benefited from the availability of landmark datasets \cite{Babenko01, Gordo01, Radenovic01, delf, Weyand01}. Apart from depicting particular landmarks, an important property of training sets~\cite{Gordo01,Radenovic01} is that they do not contain landmarks overlapping with the evaluation sets~\cite{Philbin01, Philbin02, RITAC18}. \emph{Google landmarks}~\cite{Weyand01} has gained widespread adoption in state of the art benchmarks, but falls short in this property~\cite{superfeatures}.

At the same time, a fundamental challenge in image retrieval is to find a particular object among other objects or background clutter. In this direction, it is common to use attention~\cite{Kalantidis01, Ng01, SCH01} but it is more effective use object detection~\cite{fasterrcnn, yolo} in order to represent only objects of interest for retrieval. These \emph{detect-to-retrieve} (D2R)~\cite{Teichmann01} methods however, necessitate complex two-stage training and indexing pipelines, as shown in \autoref{fig:fig1}(a), often requiring a separate training set with location supervision.

Motivated by the above challenges, we investigate two directions in this work. First, in the direction of \emph{data}, we revisit GLDv2-clean dataset~\cite{Weyand01}. We analyze and remove overlaps of landmark categories with evaluation sets~\cite{RITAC18}, introducing a new version, $\cR$GLDv2-clean. We then reproduce and benchmark state-of-the-art methods on the new dataset and compare with the original. Remarkably, we find that, although the images removed are only a tiny fraction, there is a dramatic drop in performance.

Second, in the direction of the \emph{method}, we introduce \ours, a simple attention-based approach to detect objects of interest at different levels and obtain a global image representation that effectively ignores background clutter. Importantly, as shown in \autoref{fig:fig1}(b), this is a streamlined end-to-end approach that only needs single-stage training, single-stage indexing and is free of any location supervision.

In summary, we make the following contributions:
\begin{enumerate}[itemsep=2pt, parsep=0pt, topsep=0pt]
    \item We introduce $\cR$GLDv2-clean, a new version of an established dataset for image retrieval.
    \item We show that it is critical to have no class overlap between training and evaluation sets.
    \item We introduce \ours, an end-to-end, single-stage D2R method requiring no location supervision.
    \item By using exisiting components developed outside image retrieval, we outperform more complex, specialized state-of-the-art retrieval models on several datasets.
  \end{enumerate}

\section{Related Works}
\label{sec:relatedworks}

\paragraph{Instance-level image retrieval}
Research on image retrieval can be categorized according to the descriptors used. \emph{Local descriptors}~\cite{delf, Oriane, SuperPoint} have been applied before deep learning, using SIFT~\cite{Lowe01} for example. Given that multiple descriptors are generated per image, aggregation methods~\cite{Philbin01, JPD+11, TAJ13} have been developed. Deep learning extensions include methods such as DELF~\cite{delf}, DELG~\cite{cao2020unifying}, and extensions of ASMK~\cite{Teichmann01, tolias2020learning}. DELF is similar to our work in that it uses spatial attention without location supervision, but differs in that it uses it for local descriptors.

\emph{Global descriptors}~\cite{Babenko01, Weyand01, SCH01, yang2021dolg, dtop} are useful as they only generate a single feature per image, simplifying the retrieval process.
Research has focused on spatial pooling~\cite{Razavian2015VisualIR, Radenovic01, Babenko03, Kalantidis01, ToliasSJ15, Gordo01, Radenovic01} to extract descriptors from 3D convolutional activations. Local descriptors can still be used in a second re-ranking stage after filtering by global descriptors, but this is computationally expensive.


\paragraph{Detect-to-Retrieve (D2R)}

It is beneficial for image retrieval to detect objects of interest in database images and ignore background clutter~\cite{Mei01, Salvador01, Chen01, Kucerfashion, Yining01, reddy2015object, simeoni2019graph}. Following Teichmann \etal~\cite{Teichmann01}, we call these methods \emph{detect-to-retrieve} (D2R). In most existing studies, either training or indexing are two-stage processes, for example learn to detect and learn to retrieve; also, most rely on location supervision in learning to detect.

For example, DIR~\cite{Gordo01} performs 1-stage indexing but 2-stage training for a region proposal network (RPN) and for retrieval. Its location supervision does not involve humans but rather originates in automatically analyzing the dataset, hence technically training is 3-stage. Salvador~\etal~\cite{Salvador01} performs 1-stage end-to-end training, but is using human location supervision, in fact from the \emph{evaluation set}. R-ASMK~\cite{Teichmann01}, involves 2-stage training and 2-stage indexing. It also uses large-scale human location supervision from an independent set. 

\autoref{tab:rel1} shows previous studies organized according to their properties. We can see that, unlike previous studies, we propose a novel method that supports 1-stage training, indexing and inference, as well as allowing end-to-end D2R learning without location supervision. Compared with the previous studies, ours more thus efficient.

\begin{table}
\centering
\small
\setlength{\tabcolsep}{2pt}
\begin{tabular}{lcc@{}ccccc} \toprule
	{\Th{Method}}                          & {\Th{LD}} & {\Th{GD}} &          & \Th{D2R} & \Th{E2E}  & \Th{Self} & \Th{Land} \\ \midrule
	DELF~\cite{delf}                       & \ch       &           &          &          &           &           & \ch       \\
	DELG~\cite{cao2020unifying}            & \ch       & \ch       &          &          &           &           & \ch       \\
	Tolias~\etal~\cite{tolias2020learning} & \ch       &           &          &          &           &           & \ch       \\
	DIR~\cite{Gordo01}                     &           & \ch       &          &          &           &           & \ch       \\
	AGeM~\cite{gu2018attention}            &           & \ch       &          &          &           &           & \ch       \\
	SOLAR~\cite{Ng01}                      &           & \ch       &          &          &           &           & \ch       \\
	GLAM~\cite{SCH01}                      &           & \ch       &          &          &           &           & \ch       \\ \midrule
	Kucer~\etal~\cite{Kucerfashion}        &           & \ch       &          & \ch      &           &           &           \\
	PS-Net~\cite{Yining01}                 &           & \ch       &          & \ch      &           &           &           \\
	Peng~\etal~\cite{leaf01}               &           & \ch       &          & \ch      &           &           &           \\
	Zhang~\etal~\cite{Zhang01}             &           & \ch       &          & \ch      &           &           & \ch       \\
	Liao~\etal~\cite{Liao01}               &           & \ch       &          & \ch      &           &           & \ch       \\
	R-ASMK~\cite{Teichmann01}              & \ch       &           &          & \ch      &           &           & \ch       \\
	Salvador~\etal~\cite{Salvador01}       &           & \ch       &          & \ch      & \ch       &           & \ch       \\ \midrule \rowcolor{LightCyan}
	\tb{\ours (Ours)}                      &           & \tb{\ch}  &          & \tb{\ch} & \tb{\ch}  & \tb{\ch}  & \tb{\ch}  \\ \bottomrule
\end{tabular}
\vspace{-6pt}
\caption{Related work on instance-level image retrieval. \Th{LD}: local descriptors; \Th{GD}: global descriptors. \Th{[O]}: off-the-shelf (pre-trained on ImageNet); \Th{D2R}: detect-to-retrieve; \Th{E2E} (D2R only): end-to-end (single-stage) training for detection and retrieval; \Th{Self} (D2R only): self-localization (no location supervision); \Th{Land}: landmark datasets.}
\label{tab:rel1}
\end{table}


\section{Revisiting Google Landmarks v2}
\label{sec:dataset}

\begin{figure*}
\centering
\scriptsize
\setlength{\tabcolsep}{4pt}
\begin{tabular}{ccc}
\raisebox{14pt}{\makecell{Training: \\ GLDv2-clean}} &
\fig[.4]{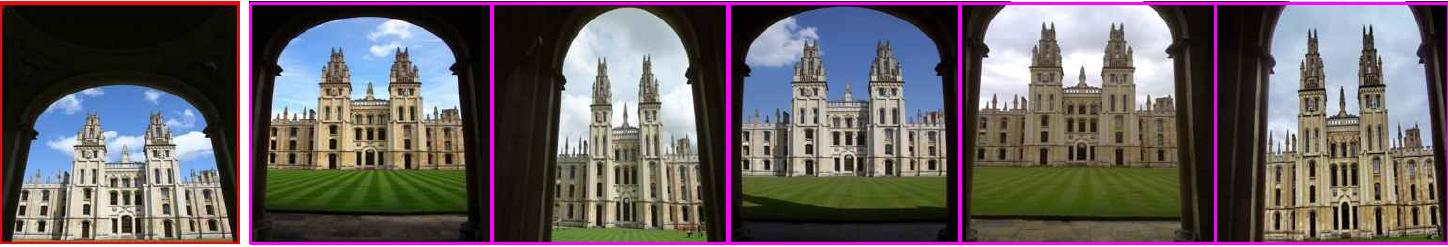} &
\fig[.4]{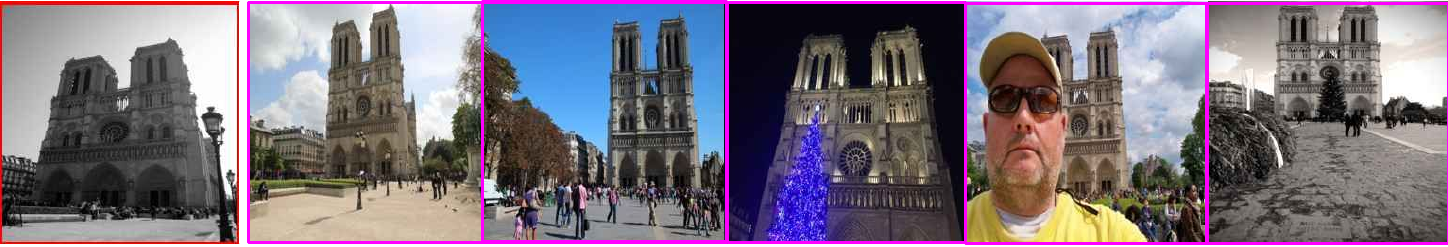} \\
\raisebox{16pt}{\makecell{Training: \\ NC-clean}} &
\fig[.4]{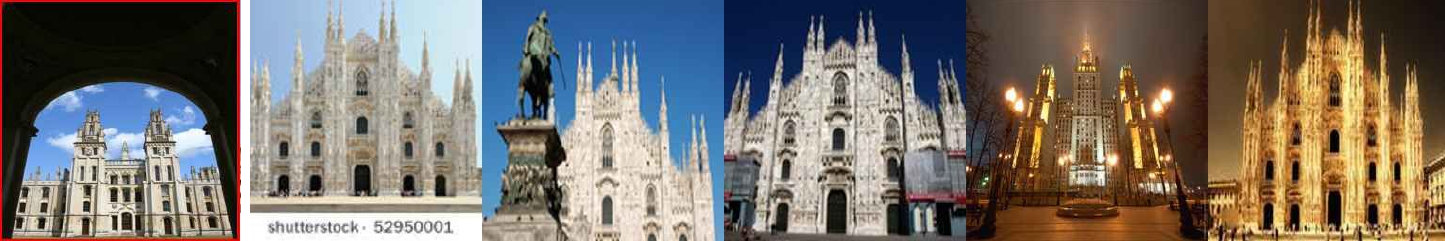} &
\fig[.4]{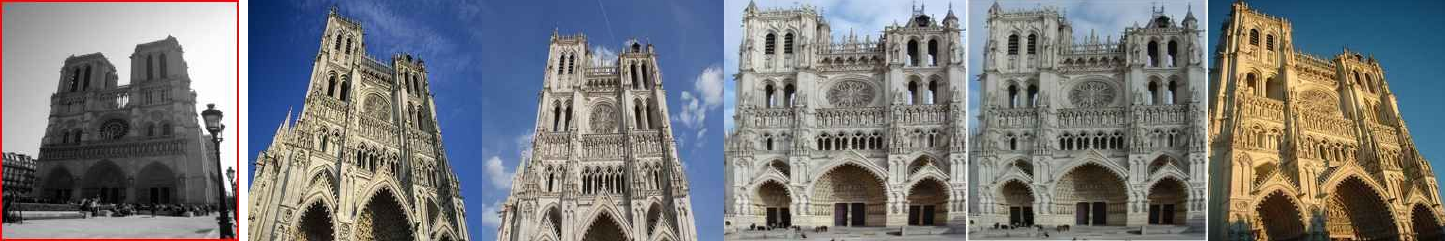} \\
\raisebox{16pt}{\makecell{Training: \\ SfM-120k}} &
\fig[.4]{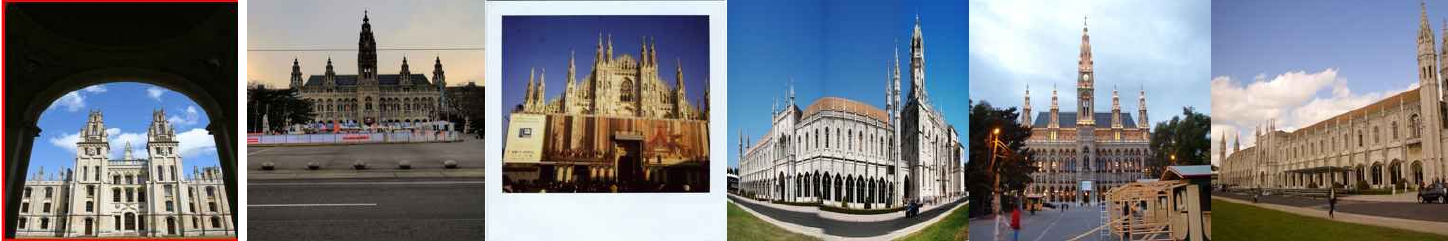} &
\fig[.4]{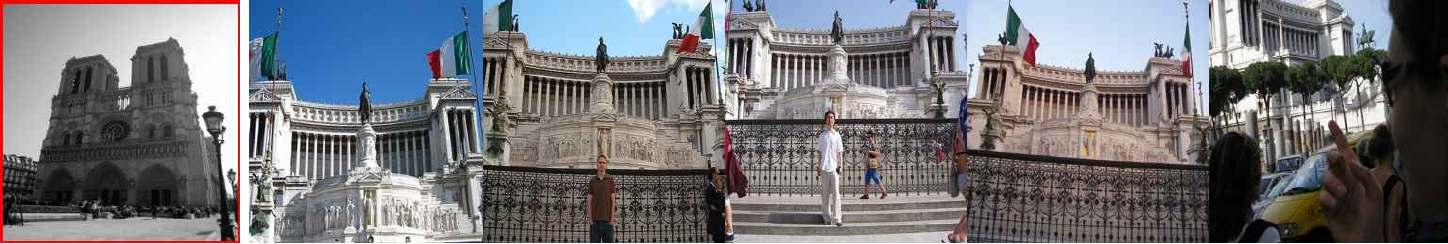} \\[3pt]
&
Evaluation: $\cR$Oxford &
Evaluation: $\cR$Paris \\
\end{tabular}
\vspace{-8pt}
\caption{\emph{Confirming overlapping landmark categories} between training sets (GLDv2-clean, NC-clean, SfM-120k) and evaluation sets ($\cR$Oxford, $\cR$Paris). Red box: query image. The query image from the evaluation set in each box/row is followed by top-5 most similar images from the training set. Pink box: training image landmark identical with query (evaluation) image landmark. More examples can be found in the Appendix.
}
\label{fig:fig2}
\end{figure*}

\paragraph{Motivation}

A key weakness of current landmark retrieval datasets is their fragmented origins: training and evaluation sets are often independently collected and released by different studies. Initial datasets contained tens of thousands of images, a number that has now grown into the millions.

\emph{Evaluation sets} such as Oxford5k (\oxf5k)~\cite{Philbin01} and Paris6k (\paris6k)~\cite{Philbin02}, as well as their more recent versions, Revisited Oxford (\roxf~or \rox) and Paris (\rpar~or \rpa)~\cite{RITAC18}, are commonly used for benchmarking. Concurrently, \emph{training sets} such as \emph{Neural Codes} (NC)~\cite{Babenko01}, \emph{Neural Codes clean} (NC-clean)~\cite{Gordo01}, SfM-120k~\cite{Radenovic01}, Google Landmarks v1 (GLDv1)~\cite{delf}, and Google Landmarks v2 (GLDv2 and GLDv2-clean)~\cite{Weyand01} have been sequentially introduced and are widely used for representation learning.

These training sets are typically curated according to two criteria: first, to depict particular landmarks, and second, to not contain landmarks that overlap with those in the evaluation sets. They are originally collected by text-based web search using particular landmark names as queries. This often results in \emph{noisy} images in addition to images depicting the landmarks. Thus, NC, GLDv1 and GLDv2 are \emph{noisy} datasets. To solve this problem, images are filtered in different ways~\cite{Gordo01, Radenovi01} to ensure that they contain only the same landmark (instance). Accordingly, NC-clean, SfM-120k, and GLDv2-clean are \emph{clean} datasets.

The \emph{clean} datasets are also typically filtered to remove overlap with the evaluation sets. However, while NC-clean and SfM-120k adhere to both criteria, GLDv2-clean falls short of the second criterion. This discrepancy is not a limitation of GLDv2-clean per se, because the dataset comes with its own split of training, index and query images. However, the community is still using the \roxf and \rpar evaluation sets, whose landmarks have not been removed from GLDv2-clean. Besides, landmarks are still overlapping between the GLDv2-clean training and index sets.

This discrepancy is particularly concerning because GLDv2-clean is the most common training set in state-of-the-art studies. It has been acknowledged in previous work~\cite{superfeatures} and in broader community discussions\footnote{\url{https://github.com/MCC-WH/Token/issues/1}}. The effect is that results of training on GLDv2-clean are not directly comparable with those of training on NC-clean or SfM-120k. Results on GLDv2-clean may show artificially \emph{inflated performance}. This is often attributed to its larger scale but may in fact be due to overlap. Our study aims to address this problem by introducing a new version of GLDv2-clean.


\paragraph{Identifying overlapping landmarks}

First, it is necessary to confirm whether common landmark categories exist between the training and evaluation sets. We extract image features from the training sets GLDv2-clean, NC-clean, and SfM-120k, as well as the evaluation sets \rox and \rpa. The features of the training sets are then indexed and the features of the evaluation sets \rox and \rpa are used as queries to search into the training sets.

\autoref{fig:fig2} displays the results. Interestingly, none of the retrieved images from NC-clean and SfM-120k training sets depict the same landmark as the query image from the evaluation set. By contrast, the top-5 most similar images from GLDv2-clean all depict the same landmark as the query. This suggests that using GLDv2-clean for training could lead to artificially \emph{inflated performance} during evaluation, when compared to NC-clean and SfM-120k. A fair comparison between training sets should require no overlap with the evaluation set.

\begin{figure}[t]
\centering
\fig[.8]{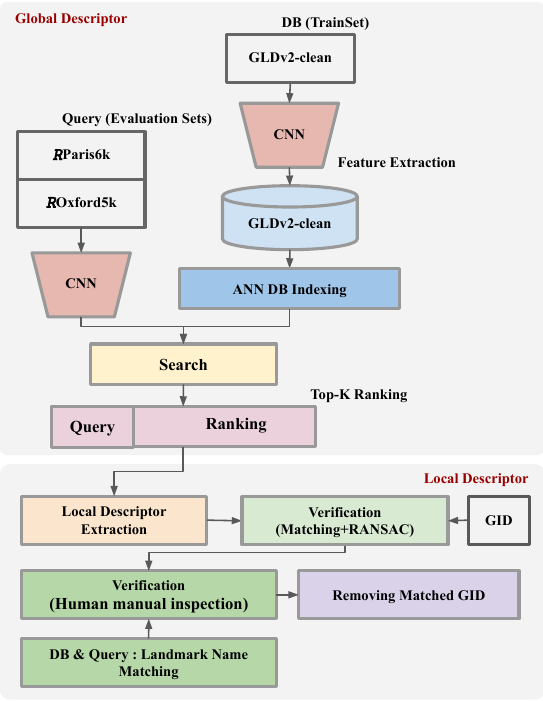}
\vspace{-6pt}
\caption{\emph{Ranking and verification pipeline} to remove landmark categories from GLDv2-clean that overlap with those of the \rox and \rpa evaluation sets and obtain the revisited version, $\cR$GLDv2-clean.}
\label{fig:figa1}
\end{figure}

\paragraph{Verification}

Now, focusing on GLDv2-clean training set, we verify the overlapping landmarks. Each image in this set belongs to a landmark category and each category is identified by a GID and has a landmark name. We begin by visual matching. In particular, we retrieve images for each query image from the evaluation set as above and we filter the top-$k$ ranked images by two verification steps.

First, we automatically verify that the same landmark is depicted by using robust spatial matching on correspondences obtained by local features and descriptors. Second, since automatic verification may fail, three human evaluators visually inspect all matches obtained in the first step. We only keep matches that are confirmed by at least one human evaluator. For every query from the evaluation set, we collect all confirmed visual matches from GLDv2-clean and we remove the entire landmark category of the GID that appears more frequently in this image collection.

Independently, we collect all GIDs where the landmark name contains ``Oxford'' or ``Paris'' and we also mark them as candidate for removal. The entire landmark category of a GID is removed if it is confirmed by at least one human evaluator that it is in one the evaluation sets. This is the case for ``Hotel des Invalides Paris''. \autoref{fig:figa1} illustrates the complete ranking and verification process.

\begin{table}
\centering
\scriptsize
\setlength{\tabcolsep}{3pt}
\begin{tabular}{lcccc} \toprule
\Th{Eval} & \Th{\#Eval Img} & \Th{\#dupl Eval} & \Th{\#dupl gldv2 GID} & \Th{\#dupl gldv2 Img} \\\midrule
\rpa  & 70 & 36 (51\%) &  11 & 1,227 \\
\rox & 70 & 38 (54\%)  &  6 & 315 \\
\Th{text} &  &  &  1  & 23 \\
\midrule
\Th{total} & 140 & 74  & 18 & 1,565 \\
\bottomrule
\end{tabular}
\vspace{-8pt}
\caption{Statistical information about duplicate images/categories with (\roxf, \rpar) and GLDV2. \Th{Eval}:Evaluation Sets. \Th{dupl}:duplicated. \Th{img}:Image. \Th{GID}:GLDV2 category. }
\label{tab:table1}
\end{table}

\begin{table}
\centering
\scriptsize
\begin{tabular}{lcc} \toprule
\Th{Training Set} & \Th{\#Images} & \Th{\#Categories} \\ \midrule
NC-clean & 27,965 & 581 \\
SfM-120k & 117,369 & 713 \\
GLDv2-clean & 1,580,470 & 81,313 \\ \midrule \rowcolor{LightCyan}
$\cR$GLDv2-clean (ours) &  1,578,905 & 81,295 \\ 
\bottomrule
\end{tabular}
\vspace{-6pt}
\caption{Statistics of clean landmark training sets for image retrieval.}
\label{tab:table2}
\end{table}

\paragraph{Revisited GLDv2-clean ($\cR$GLDv2-clean)}

By removing a number of landmark categories from GLDv2-clean as specified above, we derive a revisited version of the dataset, which we call \emph{$\cR$GLDv2-clean}. As shown in \autoref{tab:table1}, \rpa  and \rox have landmark overlap with GLDv2-clean respectively for 36 and 38 out of 70 queries, which corresponds to a percentage of 51\% and 54\%, respectively. This is a very large percentage, as it represents more than half queries in both evaluation sets. In the new dataset, we remove 1,565 images from 18 GIDs of GLDv2-clean.

\autoref{tab:table2} compares statistics between existing clean datasets and the new $\cR$GLDv2-clean. We observe that a very small proportion of images and landmark categories are removed from GLDv2-clean to derive $\cR$GLDv2-clean. Yet, it remains to find what is the effect on retrieval performance, when evaluated on \rox and \rpa. For fair comparisons, we exclude from our experiments previous results obtained by training on GLDv2-clean; we limit to NC-clean, SfM-120k and the new $\cR$GLDv2-clean.

\section{Single-stage pipeline for D2R}
\label{sec:1-stage}

\begin{figure}
\centering
\fig[1.0]{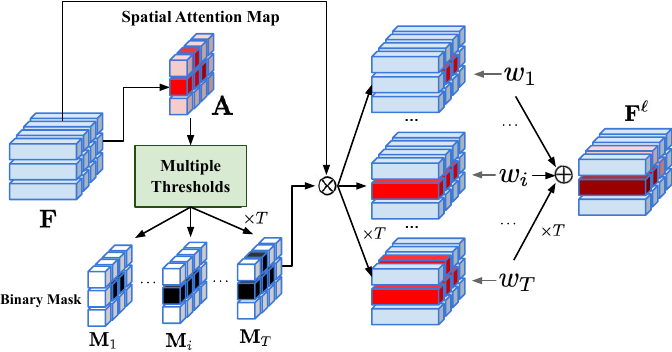}
\vspace{-16pt}
\caption{\emph{Attentional localization} (AL). Given a feature tensor $\vF \in \real^{w \times h \times d}$, we obtain a spatial attention map $A \in \real^{w \times h}$~\eq{attn} and we apply multiple thresholding operations to obtain a sequence of masks $M_1, \dots M_T$~\eq{mask}. The masks are applied independently to $\vF$ and the resulting tensors are fused into a single tensor $\vF^\ell$ by a convex combination with learnable weights $w_1, \dots, w_T$~\eq{alm}.}
\label{fig:alm}
\end{figure}

\paragraph{Motivation}

From the perspective of instance-level image retrieval, the key challenge is that target objects or instances are situated in different contexts within the image. One common solution is to use object localization or detection, isolating the objects of interest from the background. The detected objects are then used to extract an image representation for retrieval, as shown in \autoref{fig:fig1}(a). This \emph{two-stage} process can be applied to the indexed set, the queries, or both.

This approach comes with certain limitations. First, in addition to the training set for representation learning, a specialized training set is also required that is annotated with location information for the objects of interest~\cite{fasterrcnn, yolo}. Second, the two stages are often trained separately rather than end-to-end. Third, this approach incurs higher computational cost at indexing and search because it requires two forward passes through the network for each image.

In this work, we attempt to address these limitations. We replace the localization step with a \emph{spatial attention} mechanism, which does not require location supervision. This allows us to solve for both localization and representation learning through a single, end-to-end learning process on a single network, as illustrated in \autoref{fig:fig1}(b). This has the advantage of eliminating the need for a specialized training set for localization and the separate training cycles.


\paragraph{Attentional localization (AL)}

This component, depicted in \autoref{fig:fig1}(b) and elaborated in \autoref{fig:alm}, is designed for instance detection and subsequent image representation based on the detected objects. It employs a spatial attention mechanism~\cite{Kalantidis01, delf, woo01}, which does not need location supervision. Given a feature tensor $\vF \in \real^{w \times h \times d}$, where $w \times h$ is the spatial resolution and $d$ the feature dimension, we obtain the \emph{spatial attention map}
%
%
\begin{equation}
	A = \eta(\zeta(f^\ell(\vF))) \in \real^{w \times h}.
\label{eq:attn}
\end{equation}
Here, $f^\ell$ is a simple mapping, for example a $1 \times 1$ convolutional layer that reduces dimension to 1, $\zeta(x) \defn \ln(1+e^x)$ for $x \in \real$ is the softplus function and
\begin{equation}
	\eta(X) \defn \frac{X - \min X}{\max X - \min X} \in \real^{w \times h}
\label{eq:minmax}
\end{equation}
linearly normalizes $X \in \real^{w \times h}$ to the interval $[0,1]$. To identify object regions, we then apply a sequence of thresholding operations, obtaining a corresponding sequence of masks
\begin{equation}
M_i(\vp) = \left\{
	\begin{array}{ll}
		\beta,  & \mif\ A(\vp) < \tau_i \\
		1,      & \other
	\end{array}
\right.
\label{eq:mask}
\end{equation}
for $i \in \{1,\dots,T\}$. Here, $T$ is the number of masks, $\vp \in \{1,\dots,w\} \times \{1,\dots,h\}$ is the spatial position, $\tau_i \in [0,1]$ is the $i$-th threshold, $\beta$ is a scalar corresponding to background and $1$ corresponds to foreground.

Unlike a conventional fixed value like $\beta = 0$, we use a dynamic, randomized approach. In particular, for each $\vp$, we draw a sample $\epsilon$ from a normal distribution and we clip it to $[0,1]$ by defining $\beta = \min(0,\max(1,\epsilon))$. The motivation is that randomness compensates for incorrect predictions of the attention map~\eq{attn}, especially at an early stage of training. This choice is ablated in \autoref{tab:bgvalue}.

\begin{figure}
\centering
\scriptsize
\setlength{\tabcolsep}{2pt}
\begin{tabular}{cccc}
\fig[.24]{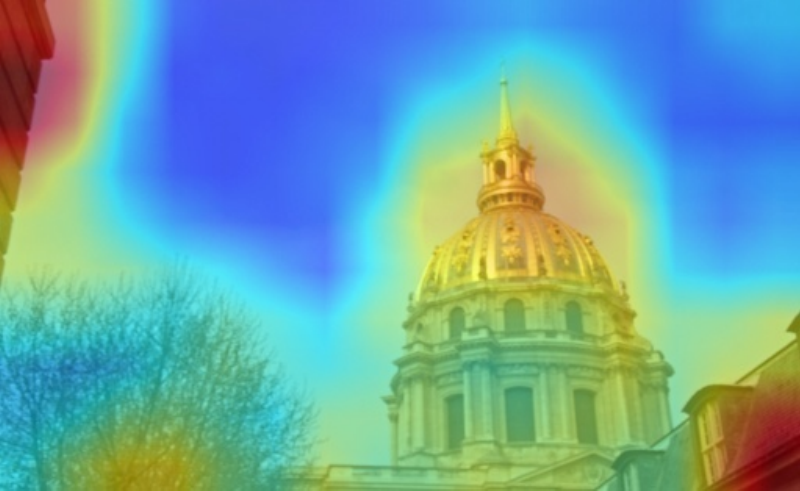} & \mc{3}{\fig[.72]{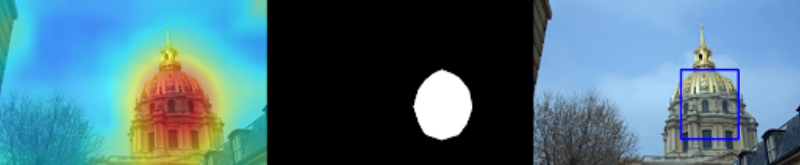}} \\[-1pt]
\fig[.24]{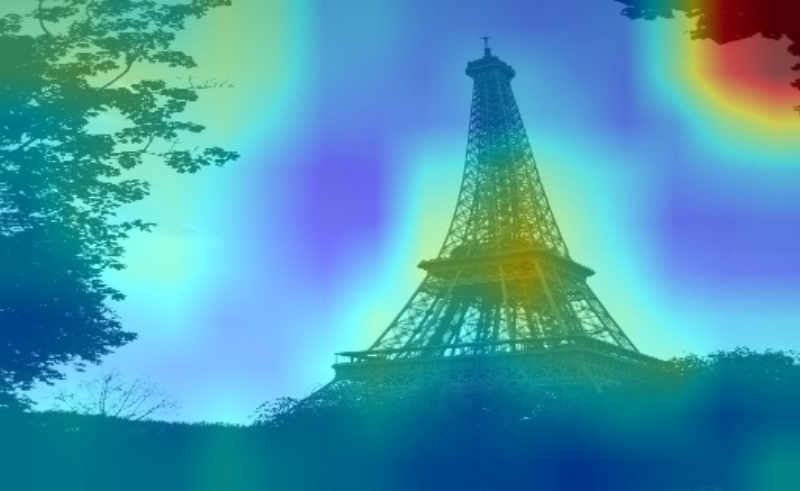} & \mc{3}{\fig[.72]{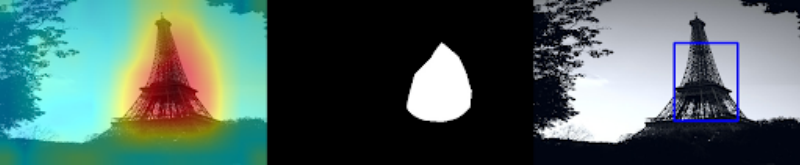}} \\[-1pt]
\fig[.24]{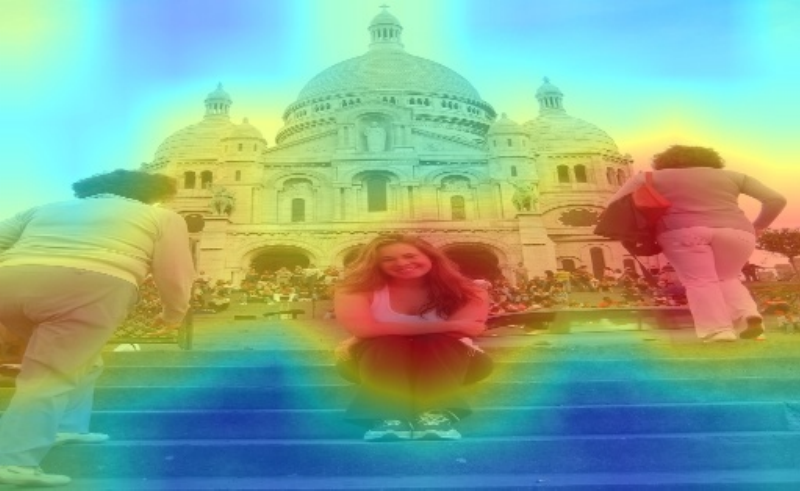} & \mc{3}{\fig[.72]{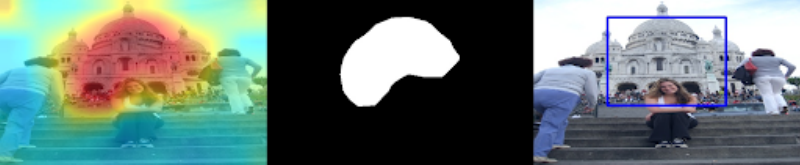}} \\[-1pt]
\fig[.24]{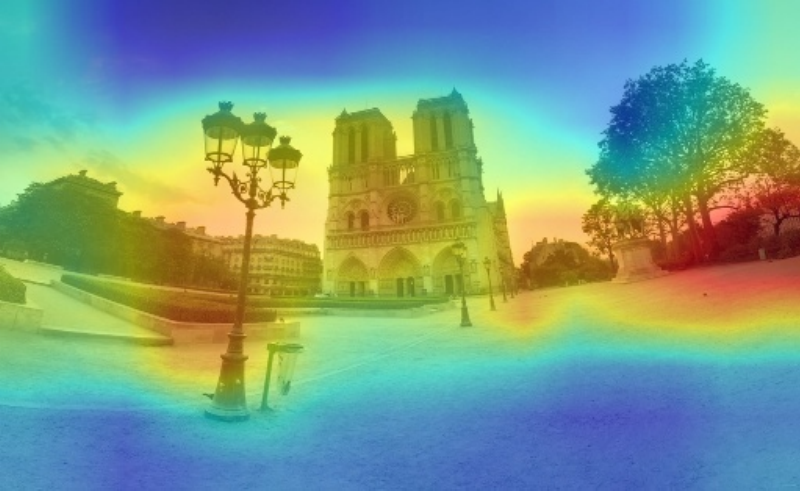} & \mc{3}{\fig[.72]{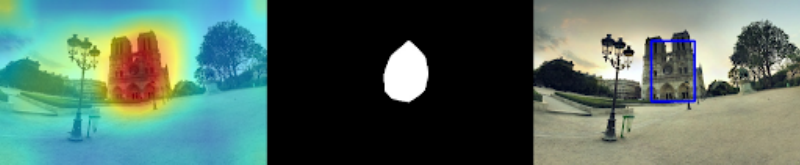}} \\[-1pt]
\fig[.24]{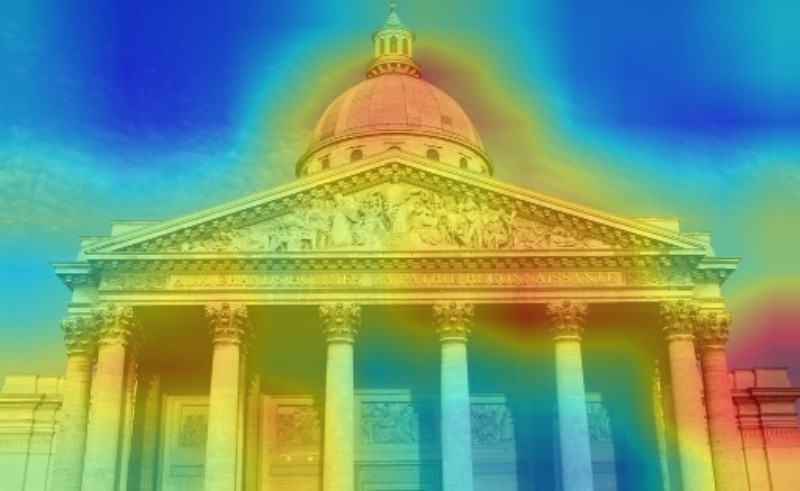} & \mc{3}{\fig[.72]{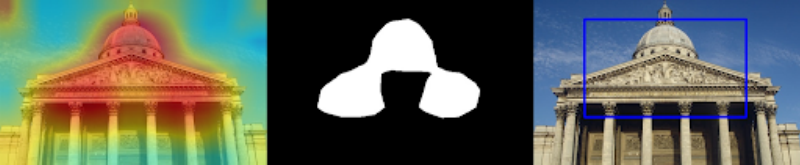}} \\
\makebox[.22\columnwidth][c]{(a) $A$, pre-trained} &
\makebox[.22\columnwidth][c]{(b) $A$, ours} &
\makebox[.22\columnwidth][c]{(c) Mask $M_i$} &
\makebox[.22\columnwidth][c]{(d) Bounding box} \\
\end{tabular}
\vspace{-6pt}
\caption{\emph{Attentional localization (AL)}. (a) Spatial attention map $A$~\eq{attn} learned on frozen ResNet101, as pre-trained on ImageNet. (b) Same, but with the network fine-tuned on $\cR$GLDv2-clean. (c) Binary mask $M_i$~\eq{mask} for $i=2$, with $\beta = 0$ for visualization. (d) Detected regions as bounding boxes of connected components of $M_i$, overlaid on input image (in blue).
}
\label{fig:fig4}
\end{figure}

\autoref{fig:fig4} shows examples of attentional localization. Comparing (a) with (b) shows that the spatial attention map generated by our model is much more attentive to the object being searched than the pretrained network. These results show that the background is removed relatively well, despite not using any location supervision at training.

The sequence of masks $M_1, \dots, M_T$~\eq{mask} is applied independently to the feature tensor $\vF$ and the resulting tensors are fused into a single tensor
%
%
\begin{equation}
	\vF^\ell = \mathtt{H}(M_1 \odot \vF, \dots, M_T \odot \vF) \in \real^{w \times h \times d},
\label{eq:alm}
\end{equation}
where $\odot$ denotes Hadamard product over spatial dimension, with broadcasting over the feature dimension. Fusion amounts to a learnable convex combination
\begin{equation}
    \mathtt{H}(\vF_1, \dots, \vF_T) = \frac{w_1 \vF_1 + \cdots + w_T \vF_T}{w_1 + \cdots + w_T},
\label{eq:fuse}
\end{equation}
where, for $i \in \{1,\dots,T\}$, the $i$-th weight is defined as $w_i = \zeta(\alpha_i)$ and $\alpha_i$ is a learnable parameter. Thus, the importance of each threshold in localizing objects from the spatial attention map is implicitly learned from data, without supervision. \autoref{tab:thresholds} ablates the effect of the number $T$ of thresholds on the fusion efficacy.


\begin{table*}
\centering
\newcolumntype{L}[1]{>{\raggedright\let\newline\\\arraybackslash\hspace{0pt}}m{#1}}
\newcolumntype{C}[1]{>{\centering\let\newline\\\arraybackslash\hspace{0pt}}m{#1}}
\newcolumntype{R}[1]{>{\raggedleft\let\newline\\\arraybackslash\hspace{0pt}}m{#1}}
\def\arraystretch{1.1}
{
\centering
\scriptsize
\setlength\extrarowheight{-1pt}
\setlength{\tabcolsep}{3pt} 
{
\begin{tabular}
{lc|cccccccccc|cc}
\toprule
\mr{3}{\Th{Method}} &	
\mr{3}{\Th{Train Set}} &	
\multicolumn{2}{c}{\Th{Base}}   & 
\multicolumn{4}{c}{\Th{Medium}} & 
\multicolumn{4}{c|}{\Th{Hard}}   &
\mr{3}{\Th{Mean}} & 
\mr{3}{\Th{Diff}} \\	\cmidrule{3-12}
& & {\fontsize{7}{6}\selectfont \oxf5k} & {\fontsize{7}{6}\selectfont \paris6k} &
\multicolumn{2}{c}{{\fontsize{7}{6}\selectfont \rox}} &
\multicolumn{2}{c}{{\fontsize{7}{6}\selectfont \rpa}} & 
\multicolumn{2}{c}{{\fontsize{7}{6}\selectfont \rox}} & 
\multicolumn{2}{c|}{{\fontsize{7}{6}\selectfont \rpa}} \\
& & \tiny mAP &  \tiny mAP &  \tiny mAP & \tiny mP@10 &  \tiny mAP & \tiny mP@10 &  \tiny mAP & \tiny mP@10 &  \tiny mAP & \tiny mP@10 \\
\midrule
    Yokoo~\etal~\cite{SCH01}  & GLDv2-clean      & 91.9 & 94.5 & 72.8 & 86.7 & 84.2 & 95.9 & 49.9 & 62.1 & 69.7 & 88.4 & 79.5 &  \gain{-5.4} \\ \rowcolor{LightCyan}
    Yokoo~\etal~\cite{Yokoo01}$^\dagger$  & $\cR$GLDv2-clean & 86.1 & 93.9 & 64.5 & 81.0 & 84.1 & 95.4 & 35.6 & 51.5 & 68.7 & 86.4 & {74.1} &  \\ \midrule
    SOLAR~\cite{tokens}       & GLDv2-clean      & --  &  --   & 79.7 & --   & 88.6 & --  & 60.0 &  --  & 75.3 & -- & 75.9 & \gain{-8} \\ \rowcolor{LightCyan}
    SOLAR~\cite{Ng01}$^\dagger$       & $\cR$GLDv2-clean & 90.6 & 94.4 & 70.8 & 84.6 & 84.1 & 95.4 & 48.0 & 62.3 & 68.7 & 86.4 & 67.9 & \\ \midrule
    GLAM~\cite{SCH01}         & GLDv2-clean      & 94.2 & 95.6 & 78.6 & 88.2 & 88.5 & 97.0 & 60.2 & 72.9 & 76.8 & 93.4 & 83.4 & \gain{-4.1}  \\ \rowcolor{LightCyan}
    GLAM~\cite{SCH01}$^\ddagger$         & $\cR$GLDv2-clean & 90.9 & 94.1 & 72.2 & 84.7 & 83.0 & 95.0 & 49.6 & 61.6 & 65.6 & 87.6 & 79.3 & \\ \midrule
    DOLG~\cite{dtop}         & GLDv2-clean      & -- & -- & 78.8 & -- & 87.8  & -- & 58.0 & -- &  74.1 & -- &   74.7 & \gain{-7.4}  \\ \rowcolor{LightCyan}
    DOLG~\cite{yang2021dolg}$^\dagger$         & $\cR$GLDv2-clean & 88.3 & 93.9 & 70.8 & 85.3 & 83.2 & 95.4 & 47.4 & 60.0 & 67.9 & 87.4 & 67.3 & \\ \midrule
    Token~\cite{tokens}       & GLDv2-clean      & --   & -- & 82.3 & -- &  75.6 &  --  & 66.6 & -- & 78.6 & -- & 75.8 & \gain{-18.2} \\ \rowcolor{LightCyan}
    Token~\cite{tokens}$^\dagger$       & $\cR$GLDv2-clean & 84.3 & 90.0 & 61.4 & 76.4 & 75.8 & 94.0 & 36.9 & 55.2 & 54.4 & 81.0 & 57.6 & \\ 
    
 \bottomrule
 \end{tabular}
}
}
\vspace{-6pt}
\caption{Comparison of the original GLDv2-clean training set with our revisited version $\cR$GLDv2-clean for a number of SOTA methods that we reproduce with ResNet101 backbone, ArcFace loss and same sampling, settings and hyperparameters. $\dagger/\ddagger$: official/our code. 
}
\label{tab:sota2_diff}
\end{table*}

\section{Experiments}
\label{sec:exp}

\subsection{Implementation}
\label{sec:impl}

\paragraph{Components}

Most instance-level image retrieval studies propose a kind of head on top of the backbone network that performs a particular operation to enhance retrieval performance. The same is happening independently in studies of category-level tasks like localization, even though the operations may be similar. Comparison is often challenging, when official code is not released. Our focus is on detection for retrieval in this work but we still need to compare with SOTA methods, which may perform different operations. We thus follow a neutral approach whereby we reuse existing, well-established components from the literature, introduced either for instance-level or category-level tasks.

In particular, given an input image $x \in \cX$, where $\cX$ is the image space, we obtain an embedding $\vu = f(x) \in \real^d$, where $d$ is the embedding dimension and
\begin{equation}
	f = f^p \circ f^\ell \circ f^c \circ f^e \circ f^b
\label{eq:map}
\end{equation}
is the composition of a number of functions. Here,
\begin{itemize}[itemsep=1pt, parsep=0pt, topsep=3pt]
	\item $f^b: \cX \to \real^{w \times h \times d}$ is the \emph{backbone network};
	\item $f^e: \real^{w \times h \times d} \to \real^{w \times h \times d}$ is \emph{backbone enhancement} (BE), including non-local interactions like ECNet~\cite{wang01}, NLNet~\cite{Wang02}, Gather-Excite~\cite{GatherExcite} or SENet~\cite{Hu01};
	\item $f^c: \real^{w \times h \times d} \to \real^{w \times h \times d}$ is \emph{selective context} (SC), enriching contextual information to apply locality more effectively like ASPP~\cite{Chen2017RethinkingAC} or SKNet~\cite{sknet};
	\item $f^\ell: \real^{w \times h \times d} \to \real^{w \times h \times d}$ is our \emph{attentional localization} (AL) (\autoref{sec:1-stage}), localizing objects of interest in an unsupervised fashion;
	\item $f^p: \real^{w \times h \times d} \to \real^d$ is a \emph{spatial pooling} operation, such as GAP or GeM~\cite{Radenovic01}, optionally followed by other mappings, \eg whitening.
\end{itemize}
In the Appendix, we ablate different options for $f^e, f^c$ and we specify our choice for $f^p$; then in \autoref{sec:ablation} we ablate, apart from hyperparameters of $f^\ell$, the effect of the presence of components $f^e, f^c, f^\ell$ on the overall performance. By default, we embed images using $f$~\eq{map}, where for each component we use default settings as specified in \autoref{sec:ablation} or in the Appendix.


\paragraph{Settings}

Certain existing works~\cite{Gordo01, delf} train the backbone network first on classification loss without the head corresponding to the method and then fine-tune including the head. We refer to this approach as ``fine-tuning'' (FT). To allow for comparisons, we train our model in two ways. \emph{Without fine-tuning}, referred to as \ours, everything is trained in a single stage end-to-end. \emph{With fine-tuning}, referred to as \oursf, we freeze the backbone while only training the head in the second stage. We give more details in the Appendix,
along with all experimental setings.

\subsection{Revisited \vs original GLDv2-clean}
\label{sec:exp-data}

We reproduce a number of state-of-the-art (SOTA) methods using official code where available, we train them on both the original GLDv2-clean dataset our revisited version $\cR$GLDv2-clean and we compare their performance on the evaluation sets. To ensure a fair evaluation, we use the same ResNet101 backbone~\cite{Gordo01, Kalantidis01, Radenovic01, gu2018attention, Ng01, Yokoo01, SCH01, yang2021dolg, tokens} and ArcFace loss~\cite{Yokoo01, SCH01, yang2021dolg, tokens, dtop} as in previous studies.

\autoref{tab:sota2_diff} shows that using $\cR$GLDv2-clean leads to severe performance degradation across all methods, ranging from 1\% up to 30\%. Because the difference between the two training sets in terms of both images and landmark categories is very small (\autoref{tab:table2}), this degradation can be safely attributed to the overlap of landmarks between the original training set, GLDv2-clean, and the evaluation sets, Oxford5k and Paris6k, as discussed in \autoref{sec:dataset}. In other words, this experiment demonstrates that existing studies using GLDv2-clean as a training set have artificially inflated accuracy metrics comparing with studies using other training sets with no overlap, such as NC-clean and SfM-120k.


\begin{table*}
\centering
\scriptsize
\setlength{\tabcolsep}{2pt}
\begin{tabular}{lcccccccc|cccccc|c} \toprule
\mr{2}{\Th{Method}} &  \mr{2}{\Th{Train Set}} & \mr{2}{\Th{Net}} & \mr{2}{\Th{Pooling}}  & \mr{2}{\Th{Loss}} & \mr{2}{\Th{FT}} &
\mr{2}{\Th{E2E}} & \mr{2}{\Th{Self}}  & \mr{2}{\Th{Dim}} &
\mc{2}{\Th{Base}} & \mc{2}{\Th{$\cR$Medium}} & \multicolumn{2}{c|}{\Th{$\cR$Hard}} & \mr{2}{\Th{Mean}} \\ \cmidrule(l){10-15}
& & & & & & & & & \Th{Oxf5k} & \Th{Par6k} &\rox & \rpa & \rox & \rpa \\

\midrule \rowcolor{lightgray2}
\multicolumn{16}{c}{\Th{Local Descriptors}} \\ \midrule
{\hesaff-\rsift-\asmk\!\!+\sp}~\cite{RITAC18}   & SfM-120k                        & R50     & --      & --           & \yes & --   & --   & --   & -- & -- & 60.6 & 61.4 & 36.7 & 35.0 & -- \\
{\delf-\asmk\!\!+\sp}~\cite{RITAC18}            & SfM-120k                        & R50     & --      & CLS          & \yes & --   & --   & --   & -- & -- & \tb{67.8} & \tb{76.9} & \tb{43.1}  & \tb{55.4} & -- \\

\midrule \rowcolor{lightgray2}
\multicolumn{16}{c}{\Th{Local Descriptors+D2R}} \\ \midrule
R-\asmk~\cite{Teichmann01}                      & NC-clean                        & R50     & --      & CLS,LOCAL    & \yes & \no  & \no  & --   & -- & -- & 69.9   & \tb{78.7} & 45.6 & \tb{57.7}   & -- \\
R-\asmk\!\!+\sp~\cite{Teichmann01}              & NC-clean                        & R50     & --      & CLS,LOCAL    & \yes & \no  & \no  & --   & -- & -- & \tb{71.9} & 78.0 & \tb{48.5} & 54.0   & -- \\

\midrule \rowcolor{lightgray2}
\multicolumn{16}{c}{\Th{Global Descriptors}} \\ \midrule
DIR~\cite{dtop}                                 & {SfM-120k}                      & R101    & RMAC    & TP           & \yes & --   & --   & 2048 & 79.0 & 86.3 & 53.5 & 68.3 & 25.5 & 42.4 &   59.2 \\
Radenovic~\etal~\cite{Radenovic01, RITAC18}     & {SfM-120k}                      & R101    & GeM     & SIA          & \no  & --   & --   & 2048 &  87.8 & \tb{92.7} & 64.7 & 77.2 & 38.5 & 56.3 &  69.5 \\
AGeM~\cite{gu2018attention}                     & {SfM-120k}                      & R101    & GeM     & SIA          & \no  & --   & --   & 2048 &  -- & -- & \tb{67.0} & \tb{78.1} & \tb{40.7} & \tb{57.3} & --  \\
SOLAR~\cite{dtop}                               & {SfM-120k}                      & R101    & GeM     & TP,SOS       & \yes & --   & --   & 2048 & 78.5 & 86.3 & 52.5 & 70.9  & 27.1 & 46.7 & 60.3 \\
GLAM~\cite{SCH01}                               & {SfM-120k}                      & R101    & GeM     & AF           & \no  & --   & --   & 512  & \tb{89.7} & 91.1 & 66.2 & 77.5 &  39.5 & 54.3 & \tb{69.7} \\
DOLG~\cite{dtop}                                & {SfM-120k}                      & R101    & GeM,GAP & AF           & \no  & --   & --   & 512  & 72.8 & 74.5 & 46.4 & 56.6 & 18.1 & 26.6 & 49.2 \\

\midrule \rowcolor{lightgray2}
\multicolumn{16}{c}{\Th{Global Descriptors+D2R}} \\ \midrule
Mei~\etal~\cite{Mei01}                          & [O]                             & R101    & FC      & CLS          & \no  & \no  & \no  & 4096 & 38.4 & --   & --  & --  &  --  & --   & -- \\
Salvador~\etal~\cite{Salvador01}                & Pascal VOC                      & V16     & GSP     & CLS,LOCAL    & \no  & \yes  & \no  & 512  & 67.9 & 72.9   & --  & --  &  --  & --   & -- \\
Chen~\etal~\cite{Chen01}                        & OpenImageV4~\cite{Kuznetsova01} & R50     & MAC     & MSE          & \no  & \yes & \no  & 2048 & 50.2 & 65.2 & --  & --  &  --  & --   & -- \\
Liao~\etal~\cite{Liao01}                        & Oxford,Paris                    & A,V16   & CroW    & CLS,LOCAL    & \no  & \no  & \no  & 768  & 80.1 & 90.3 & --  & --  &  --  & --   & -- \\
DIR+RPN~\cite{Gordo01}                          & {NC-clean}                      & R101    & RMAC    & TP           & \yes & \no  & \no  & 2048 & \tb{85.2} & \tb{94.0} & --  & --  &  --  & --   & -- \\

\midrule \rowcolor{LightCyan}
\tb{\ours (Ours)}                               & {SfM-120k}                      & R101    & GeM     & AF           & \no  & \yes & \yes & 2048 & \ok{\tb{89.9}} & {92.0} & \ok{\tb{67.3}} & \ok{\tb{79.4}} & \ok{\tb{42.4}} & {{57.5}} &  \ok{\tb{71.4}} \\ \rowcolor{LightCyan}
\tb{\oursf (Ours)}                              & {SfM-120k}                      & R101    & GeM     & AF           & \yes & \yes & \yes & 2048 & \red{\tb{92.6}} & \red{\tb{95.1}} & \red{\tb{76.2}}&  \red{\tb{84.5}}& \red{\tb{58.9}} & \red{\tb{68.9}} &  \red{\tb{79.4}}\\

\bottomrule
\end{tabular}

\vspace{-8pt}
\caption{Properties and mAP comparison of SOTA on existing training sets with no overlap with evaluation sets.
\Th{FT}: fine-tuning; \Th{E2E} (D2R only): end-to-end (single-stage) training for detection and retrieval; \Th{Self} (D2R only): self-localization (no location supervision).
\emph{Network}: R50/101: ResNet50/101; V16: VGG16; A: AlexNet.
\emph{Pooling}: GAP: global average pooling; GSP: global sum pooling.
\emph{Loss}: AF: ArcFace; TP: triplet; CLS: softmax; SIA: siamese; SOS: second-order similarity; MSE: mean square error; LOCAL: Localization Loss;
SP: spatial verification.
[O]: Off-the-shelf (pre-trained on ImageNet).
Red: best result; blue: our results higher than previous methods; black bold: best previous method per block.
}
\label{tab:sota_cleandata}
\end{table*}

\begin{table*}
\newcolumntype{L}[1]{>{\raggedright\let\newline\\\arraybackslash\hspace{0pt}}m{#1}}
\newcolumntype{C}[1]{>{\centering\let\newline\\\arraybackslash\hspace{0pt}}m{#1}}
\newcolumntype{R}[1]{>{\raggedleft\let\newline\\\arraybackslash\hspace{0pt}}m{#1}}
\def\arraystretch{1.13}
\centering
\scriptsize
\setlength\extrarowheight{-1pt}
\setlength{\tabcolsep}{2pt}
{
\begin{tabular}
{l|cc|cccccccc|cccccccc}
	\toprule
	\multirow{3}{*}{\Th{Method}} &
	\multicolumn{2}{c|}{\Th{Base}}  & \multicolumn{8}{c|}{\Th{Medium}} & \multicolumn{8}{c}{\Th{Hard}} \\ \cmidrule{2-19}

	&  {\fontsize{7}{6}\selectfont \oxf5k} & {\fontsize{7}{6}\selectfont \paris6k} & \multicolumn{2}{c}{{\fontsize{7}{6}\selectfont \rox}} & \multicolumn{2}{c}{{\fontsize{7}{6}\selectfont \rox+\r1m}} & \multicolumn{2}{c}{{\fontsize{7}{6}\selectfont \rpa}} & \multicolumn{2}{c|}{{\fontsize{7}{6}\selectfont \rpa+\r1m}} & \multicolumn{2}{c}{{\fontsize{7}{6}\selectfont \rox}} & \multicolumn{2}{c}{{\fontsize{7}{6}\selectfont \rox+\r1m}} & \multicolumn{2}{c}{{\fontsize{7}{6}\selectfont \rpa}} & \multicolumn{2}{c}{{\fontsize{7}{6}\selectfont \rpa+\r1m}} \\

	& \tiny mAP &  \tiny mAP & \tiny mAP & \tiny mP@10 & \tiny mAP & \tiny mP@10 & \tiny mAP & \tiny mP@10 & \tiny mAP & \tiny mP@10 & \tiny mAP & \tiny mP@10 & \tiny mAP & \tiny mP@10 & \tiny mAP & \tiny mP@10 & \tiny mAP & \tiny mP@10 \\
	\midrule
	\rowcolor{lightgray2}
	\multicolumn{19}{c}{\Th{Global Descriptors (SfM-120k)}} \\ \midrule
	DIR~\cite{dtop}  & 79.0 & 86.3 & 53.5 & 76.9& -- & -- & 68.3 & 97.7 & -- & -- & 25.5 & 42.0 & -- & --& 42.4 & 83.6 & -- & -- \\
	%
	%
    %
	Filip~\etal~\cite{Radenovic01,RITAC18}   & 87.8 & {92.7} & 64.7 & \tb{84.7} & \tb{45.2} & \tb{71.7} & 77.2 & \tb{98.1} & \tb{52.3} & \tb{95.3} & 38.5 & \tb{53.0} & \tb{19.9} & \tb{34.9} & 56.3 & \tb{89.1} & 24.7 & \tb{73.3} \\
	AGeM~\cite{gu2018attention}   &  -- & -- & \tb{67.0} & -- & -- & -- & \tb{78.1} & --  & -- & -- & \tb{40.7} & -- & -- & -- & 57.3 & --  & -- & --  \\
	SOLAR~\cite{dtop}       & 78.5 & 86.3 & 52.5 & 73.6 & -- & -- & 70.9 & \tb{98.1} & -- & -- & 27.1 & 41.4 & -- & -- & 46.7 & 83.6 & -- & -- \\
	GeM~\cite{dtop}  & 79.0 & 82.6 & 54.0 & 72.5 & -- & -- & 64.3 & 92.6 & -- & -- & 25.8 & 42.2 & -- & -- & 36.6 & 67.6 & -- & -- \\
	GLAM~\cite{dtop}  & \tb{89.7} & 91.1 & 66.2 & -- & -- & -- & 77.5 & -- & -- & -- & 39.5 & -- & -- & -- & 54.3 & -- & -- & -- \\
	DOLG~\cite{dtop}  & 72.8 & 74.5 & 46.4 & 66.8 & -- & -- & 56.6 & 91.1 & -- & -- & 18.1 & 27.9 & -- & -- & 26.6 & 62.6 & -- & -- \\
 \midrule \rowcolor{LightCyan}
\tb{\ours (Ours)} & \ok{\tb{89.9}} & {92.0} & \ok{\tb{67.3}}& \ok{\tb{85.1}} & \ok{\tb{50.3}} & \ok{\tb{75.5}} &  \ok{\tb{79.4}}& 97.9 & 51.4 & \ok{\tb{95.7}} & \ok{\tb{42.4}} & \ok{\tb{56.4}} & \ok{\tb{22.4}} & \ok{\tb{35.9}} & {{57.5}} & 87.1 & 22.4 & 69.4 \\ \rowcolor{LightCyan}
\tb{\oursf (Ours)} & \red{\tb{92.6}} & \red{\tb{95.1}} & \red{\tb{76.2}}& \red{\tb{87.3}} & \red{\tb{60.5}} & \red{\tb{78.6}} &  \red{\tb{84.5}} & 98.0 & \red{\tb{56.9}} & \red{\tb{95.9}} & \red{\tb{58.9}} & \red{\tb{71.1}} & \red{\tb{36.8}} & \red{\tb{55.7}} & \red{\tb{68.9}} & \red{\tb{91.3}} & \red{\tb{30.1}} & \red{\tb{73.9}} \\ \rowcolor{LightCyan}
        \midrule
	\rowcolor{lightgray2}
	\multicolumn{19}{c}{\Th{Global Descriptors ($\cR$GLDV2-clean)}} \\ \midrule
        Yokoo~\etal~\cite{Yokoo01}$^\dagger$ (Base)    & 86.1   & 93.9  & 64.5 & 81.0 & 51.3 & 72.1 & 84.1 & \tb{95.4} & 54.2 & 90.3 & 35.6 & 51.5 & 22.2 & 42.9 & \tb{68.7} & 86.4 & 27.4 & 66.9   \\
        SOLAR~\cite{Ng01}$^\dagger$   & 90.6 & \tb{94.4} & 70.8 & 84.6 & 55.8 & 76.1 & 80.3 & 94.6 & 57.6 & \tb{92.0} & 48.0 & \tb{62.3} & 30.3 & 45.3 & 61.8 & 83.9 & 30.7 & 71.6  \\
        GLAM~\cite{SCH01}$^\ddagger$   & \tb{90.9} & 94.1 & \tb{72.2} & 84.7 & \tb{58.6} & 76.1 & 83.0 & 95.0  & \tb{58.6} & 91.7 & \tb{49.6} & 61.6 & \tb{34.1} & \tb{50.9} & 65.6 & \tb{87.6}  & \tb{33.3} & 72.1  \\
        DOLG~\cite{yang2021dolg}$^\dagger$        & 88.3      & 93.9     & 70.8 & \tb{85.3} & 57.3  & \tb{76.8} & \tb{83.2} & \tb{95.4} & 57.3 & \tb{92.0} &47.4 & 60.0 & 29.5 & 46.2 & 67.9 & 87.4 & 32.7 & \tb{72.4} \\
        Token~\cite{tokens}$^\dagger$             & 81.2      & 89.6     & 60.8 & 77.7 & 44.0 & 60.9 & 75.8 & 94.3 &  44.1 & 86.9 & 37.3 & 54.1 & 23.2 & 37.7 & 54.8 & 81.3 & 19.7 & 54.4 \\
        \midrule \rowcolor{LightCyan}
        \tb{\ours (Ours)}   & 89.8 & \ok{\tb{94.6}} & \ok{\tb{73.7}} & \ok{\tb{85.5}} & \ok{\tb{58.6}} & 76.3 & \ok{\tb{84.6}} & \ok{\tb{96.7}} & \ok{\tb{59.0}} & \red{\tb{95.1}} & \ok{\tb{54.9}} & \ok{\tb{66.6}} & \ok{\tb{34.6}} & \ok{\tb{54.7}} &\ok{\tb{68.5}} & \ok{\tb{89.1}} & \ok{\tb{33.5}} & \red{\tb{76.9}} \\ \rowcolor{LightCyan}
        \tb{\oursf (Ours)}   & \red{\tb{90.9}} & \red{\tb{96.1}} & \red{\tb{77.8}} & \red{\tb{88.0}} & \red{\tb{61.8}} & \red{\tb{78.0}} &\red{\tb{87.4}} & \red{\tb{97.0}} & \red{\tb{61.6}} & \ok{\tb{94.3}} & \red{\tb{61.9}} & \red{\tb{70.4}} & \red{\tb{39.4}} & \red{\tb{56.8}} &\red{\tb{75.3}} & \red{\tb{90.0}} & \red{\tb{35.8}} & \ok{\tb{72.7}} \\
	\bottomrule
\end{tabular}
}

\vspace{-8pt}
\caption{Large-scale mAP comparison of SOTA on training sets with no overlap with evaluation sets. In the new $\cR$GLDv2-clean, settings are same as in \autoref{tab:sota2_diff}. In the existing SfM-120k, results are as published. $\dagger/\ddagger$: official/our code. Red: best results; blue: our results higher than previous methods; black bold: best previous method per block. \Th{FT}:fine-tuning.}
\label{tab:1m}
\end{table*}

\subsection{Comparison with state of the art}
\label{sec:exp-bench}

\paragraph{Existing clean datasets}

\autoref{tab:sota_cleandata} compares different methods using global or local descriptors, with or without a D2R approach, on existing \emph{clean datasets} NC-clean and SfM-120k, which do not overlap with the evaluation sets.

Comparing with methods using global descriptors without D2R, our method demonstrates SOTA performance and brings significant improvements over AGeM~\cite{gu2018attention}, the previous best competitor. In particular, 2.9\%, 0.6\% mAP on \oxf5k, \paris6k Base, 9.2\%, 18.2\% on \rox, \rpa Medium, and 6.4\%, 9.5\% on \rox, \rpa Hard.



Comparing with methods using global descriptors without D2R, our method outperforms the highest-ranking approach by DIR+RPN~\cite{Gordo01}, which was trained on the SfM-120k dataset. Specifically, our method improves mAP by 7.4\% on \oxf5k dataset and by 1.1\% on \paris6k.
Interestingly, methods in the D2R category employ different training sets, as no single dataset provides annotations for both D2R tasks. Our study is unique in being single-stage, end-to-end (\textsc{E2E}) trainable and at the same time requiring no location supervision (\textsc{Loc}), thereby eliminating the need for a detection-specific training set.


\paragraph{New clean dataset, distractors}

\autoref{tab:1m} provides complete experimental results, including the impact of introducing 1 million distractors (\r 1m) into the evaluation set, on our new clean training set, $\cR$GLDv2-clean, as well as the previous most popular clean set, SfM-120k. Contrary to previous studies, we compare methods trained on the same training and evaluation sets to ensure fairness.

Without fine-tuning, we improve 1.3\% mAP on \rox$\!+$\r1m~(medium), 5.1\% on \rox$\!+$\r1m~(hard), 1.7\% on \rpar$\!+$\r1m~(medium), and 0.8\% on \rpar$\!+$\r1m~(hard) compared to DOLG~\cite{yang2021dolg} on $\cR$GLDv2-clean. With fine-tuning, our \oursf establishes new SOTA for nearly all metrics. In particular, we improve 4.5\% mAP on \rox$\!+$\r1m~(medium), 5.3\% on \rox$\!\!+$\r1m~(hard), 4.3\% on \rpar$\!+$\r1m~(medium), and 3.1\% on \rpar$\!+$\r1m~(hard) compared to DOLG~\cite{yang2021dolg} on $\cR$GLDv2-clean.

\subsection{Visualization}
\label{sec:Visualization}

\begin{figure*}
\centering
\scriptsize
\setlength{\tabcolsep}{1pt}
\begin{tabular}{ccccccccccccc}
\figwh{.07}{.06}{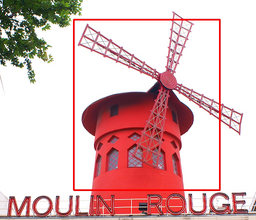} &
\figwh{.07}{.06}{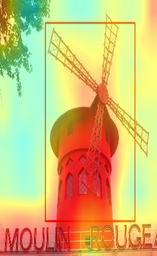} & \hspace{6pt} &
\figwh{.07}{.06}{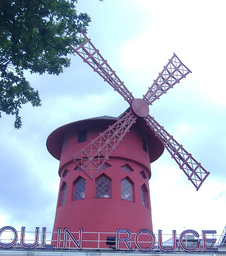} &
\figwh{.07}{.06}{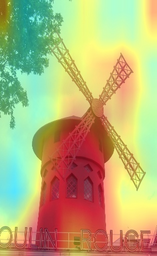} &
\figwh{.07}{.06}{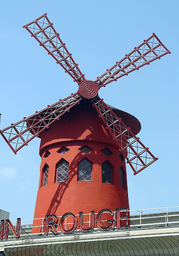} &
\figwh{.07}{.06}{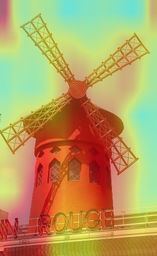} &
\figwh{.07}{.06}{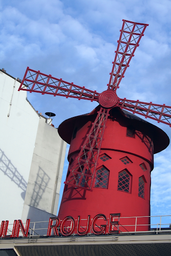} &
\figwh{.07}{.06}{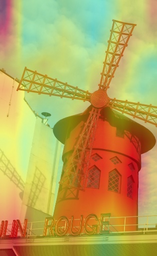} &
\figwh{.07}{.06}{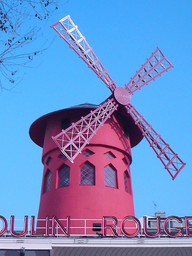} &
\figwh{.07}{.06}{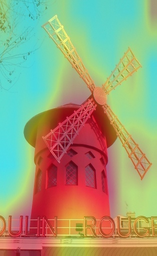} &
\figwh{.07}{.06}{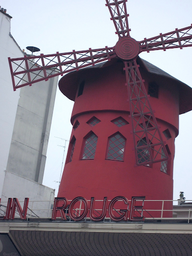} &
\figwh{.07}{.06}{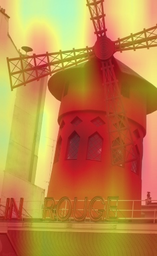} \\
\figwh{.07}{.06}{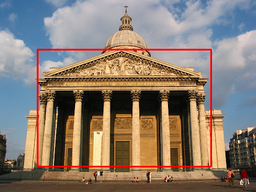} &
\figwh{.07}{.06}{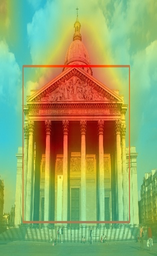} & \hspace{6pt} &
\figwh{.07}{.06}{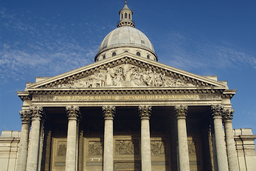} &
\figwh{.07}{.06}{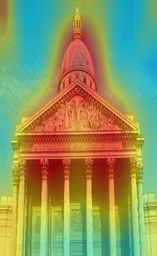} &
\figwh{.07}{.06}{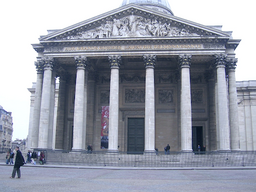} &
\figwh{.07}{.06}{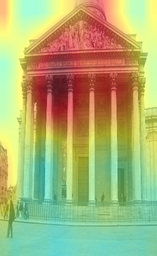} &
\figwh{.07}{.06}{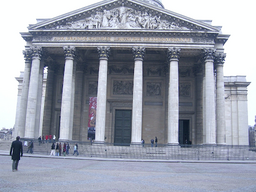} &
\figwh{.07}{.06}{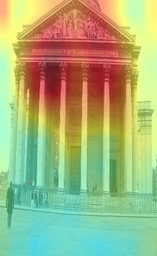} &
\figwh{.07}{.06}{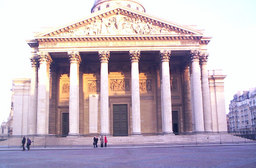} &
\figwh{.07}{.06}{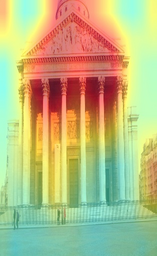} &
\figwh{.07}{.06}{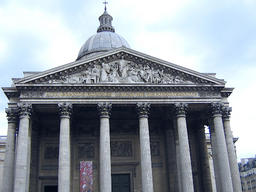} &
\figwh{.07}{.06}{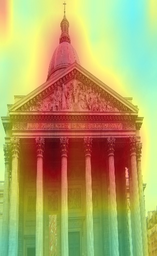} \\
\figwh{.07}{.09}{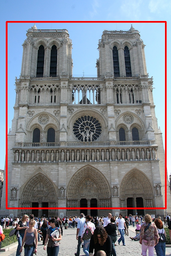} &
\figwh{.07}{.09}{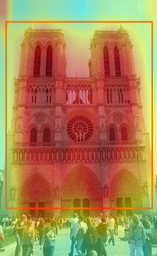} & \hspace{6pt} &
\figwh{.07}{.09}{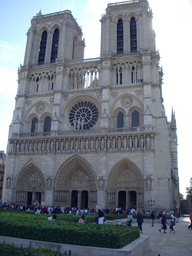} &
\figwh{.07}{.09}{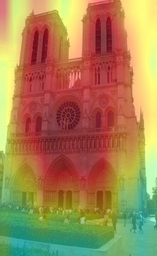} &
\figwh{.07}{.09}{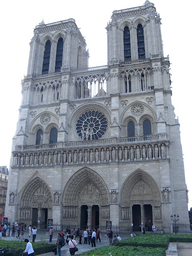} &
\figwh{.07}{.09}{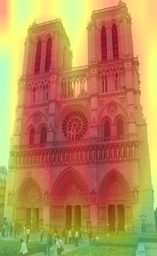} &
\figwh{.07}{.09}{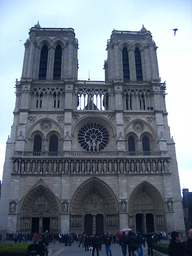} &
\figwh{.07}{.09}{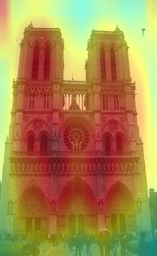} &
\figwh{.07}{.09}{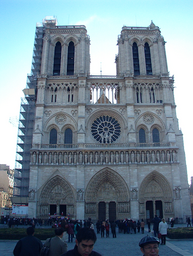} &
\figwh{.07}{.09}{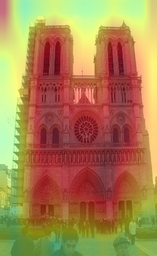} &
\figwh{.07}{.09}{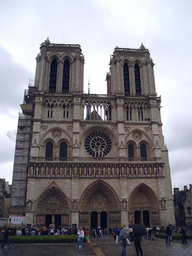} &
\figwh{.07}{.09}{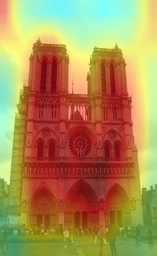} \\
Query & & &
Top-1 & &
Top-2 & &
Top-3 & &
Top-4 & &
Top-5 & \\
\end{tabular}
\vspace{-8pt}
\caption{Examples of top-5 ranking images retrieved by our \ours model from evaluation sets \oxf5k/\paris6k and associated spatial attention map $A$~\eq{attn}. The red rectangle within the query on the left is the cropped area provided by the evaluation set and is actually used as the query image.}
\label{fig:fig5}
\end{figure*}

\begin{figure}
\centering
\scriptsize
\setlength{\tabcolsep}{1pt}
\begin{tabular}{ccc}
& Pre-trained & Ours \\
\raisebox{.14\columnwidth}{(a)} & \fig[.4]{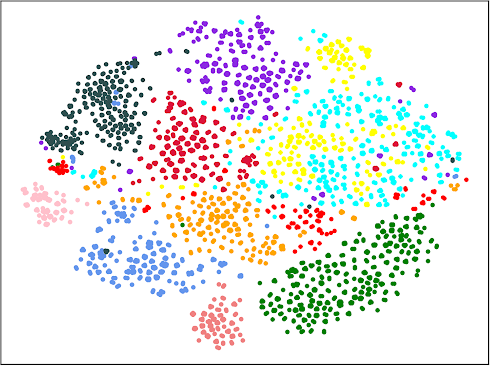} & \fig[.4]{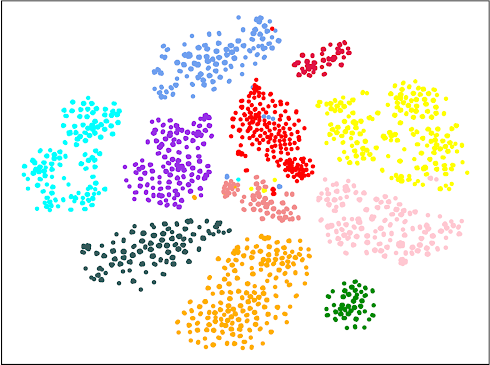} \\[-1pt]
\raisebox{.14\columnwidth}{(b)} & \fig[.4]{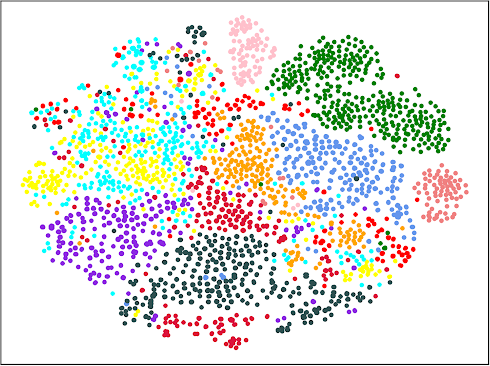} & \fig[.4]{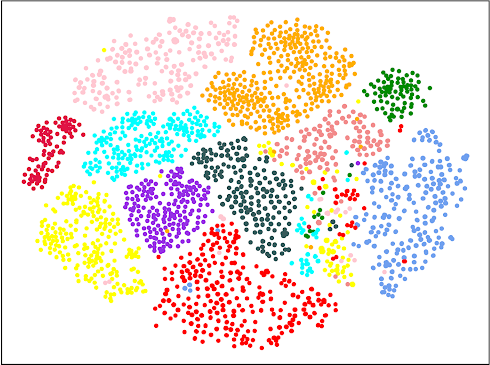} \\[-1pt]
\raisebox{.14\columnwidth}{(c)} & \fig[.4]{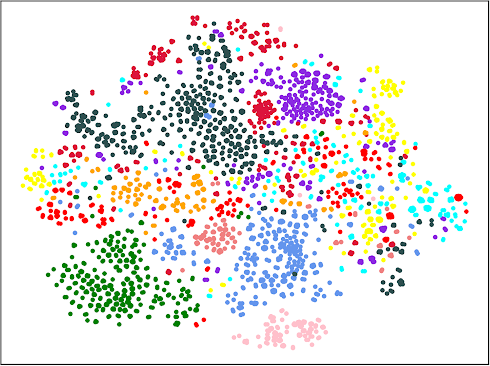} & \fig[.4]{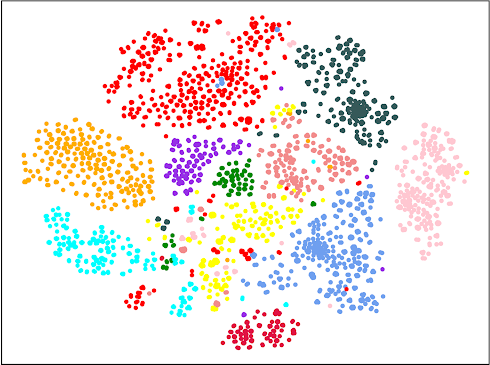}
\end{tabular}
\vspace{-9pt}
\caption{T-SNE visualization of image embeddings of the \emph{revisited Paris} (\rpa) evaluation set under (a) \emph{easy}, (b) \emph{medium}, and (c) \emph{hard} protocols~\cite{RITAC18}. Pre-trained: ResNet101 off-the shelf as pre-trained on ImageNet. Ours: our \oursf with fine-tuning on SfM-120k~\cite{Radenovic01}. Positive images for each protocol are colored based on their query landmark category. 
}
\label{fig:fig5_embedding}
\end{figure}

\paragraph{Ranking and spatial attention}

\autoref{fig:fig5} shows examples of the top-5 ranking images retrieved for a number of queries by our model, along with the associated spatial attention map.
The spatial attention map $A$~\eq{attn} focuses exclusively on the object of interest as specified by the cropped area provided by the evaluation set, essentially ignoring the background.


\paragraph{Embedding space}

\autoref{fig:fig5_embedding} shows t-SNE visualizations of image embeddings of the \rpar dataset~\cite{RITAC18}, obtained by the off-the shelf network as pre-trained on ImageNet \vs our method with fine-tuning on SfM-120k~\cite{Radenovic01}. It indicates superior embedding quality for our model. 

\subsection{Ablation study}
\label{sec:ablation}



\paragraph{Design ablation}

We study the effect of the presence of components $f^e, f^c, f^\ell$~\eq{map} on the overall performance of the proposed model. Starting from the baseline, which is ResNet101 backbone ($f^b$) followed by GeM pooling ($f^p$), we add selective context (SC, $f^c$), attentional localization (AL, $f^\ell$) and backbone enhancement (BE, $f^e$). \autoref{tab:components} provides the results, illustrating the performance gains achieved by the proposed components.


\begin{table}
\centering
\scriptsize
\setlength{\tabcolsep}{3.0pt}
\begin{tabular}{ccccccccc} \toprule
\mr{2}{SC} & \mr{2}{AL} & \mr{2}{BE} & \mr{2}{\Th{Oxf5k}} & \mr{2}{\Th{Par6k}} & \mc{2}{\Th{Medium}} & \mc{2}{\Th{Hard}} \\ \cmidrule(l){6-9}
& & & &  & \rox & \rpa & \rox & \rpa  \\ \midrule
    &     &      &   80.2    & 83.2       & 55.1       & 67.7     & 25.8      & 40.7 \\
\ch &     &      &   87.6    & 90.7       & 64.7       & 76.6      & 38.2      & 52.7 \\
\ch &     & \ch  &   89.4    & 91.1       & 66.1       & 76.7      & 40.6      & 53.3 \\
    & \ch & \ch  &   88.2    & 91.5       & 66.0       & 78.4      & 40.8      & 55.9 \\
\ch & \ch &      &   89.7    & \tb{92.0}  & 67.0       & \tb{79.4} & 41.0      & 57.4 \\
\ch & \ch & \ch  & \tb{89.9} & \tb{92.0} & \tb{67.3} & \tb{79.4} & \tb{42.4} & \tb{57.5} \\
\bottomrule
\end{tabular}
\vspace{-7pt}
\caption{Effect of different components on mAP performance. Training on SfM-120k. Baseline: ResNet101 with GeM pooling. SC: selective context; AL: attentional localization; BE: backbone enhancement. }

\label{tab:components}
\end{table}

\paragraph{Mask background $\beta$}

We study the effect of setting the background value $\beta$ in masks~\eq{mask} to a fixed value \vs clipping a sample $\epsilon$ from the normal distribution. \autoref{tab:bgvalue} indicates that our dynamic, randomized approach is superior when $\epsilon \sim \cN(0.1, 0.9)$, which we choose as default.

\begin{table}
\centering
\scriptsize
\setlength{\tabcolsep}{2pt}
\begin{tabular}{l*{7}{c}} \toprule
\mr{2}{\Th{$\beta$ setting}} & \mr{2}{\Th{Oxf5k}} & \mr{2}{\Th{Par6k}} & \mc{2}{\Th{Medium}} & \mc{2}{\Th{Hard}} \\ \cmidrule(l){4-7}
 & & &  \rox & \rpa & \rox & \rpa \\ \midrule
Fixed (0.0) & 87.4 & 91.6 & 64.9  & 77.5 & 39.1 & 53.8  \\
Fixed (0.5) & 87.5 & 91.7 & 64.8 & 77.7 & 38.8 & 54.3  \\
$\cN$(0.1, 0.5) & \tb{90.2} & 90.5 & \tb{67.4} & 78.1 & 40.2 & 55.2  \\ \rowcolor{LightCyan}
$\cN$(0.1, 0.9) & {89.9} & \tb{92.0} & {{67.3}}&  {\tb{79.4}}& {\tb{42.4}} & {\tb{57.5}}\\ \bottomrule
\end{tabular}
\vspace{-8pt}
\caption{Effect on mAP of different \emph{mask background} $\beta$~\eq{mask} settings in our attentional localization. Training on SfM-120k.}
\label{tab:bgvalue}
\end{table}

\paragraph{Number of masks $T$}

We study the effect of the number of masks $T$~\eq{mask} in our attentional localization, obtained by thresholding operations on the spatial attention map $A$~\eq{attn}. \autoref{tab:thresholds} shows that optimal performance is achieved for $T=2$, which we choose as default.

\begin{table}
\centering
\scriptsize
\setlength{\tabcolsep}{3.0pt}
\begin{tabular}{l*{7}{c}} \toprule
\mr{2}{$T$} & \mr{2}{\Th{Oxf5k}} & \mr{2}{\Th{Par6k}} & \mc{2}{\Th{Medium}} & \mc{2}{\Th{Hard}} \\ \cmidrule(l){4-7}
 & & &  \rox & \rpa & \rox & \rpa \\ \midrule
1 & {87.5} & {91.7} & {64.8}&  {77.7}& {38.8} & {54.3}\\ \rowcolor{LightCyan}
2 & \tb{89.9} & {92.0} & {{67.3}}&  {\tb{79.4}}& {\tb{42.4}} & {\tb{57.5}} \\
3 & 89.4 & \tb{92.2} & \tb{67.5} & 78.5 & \tb{42.4} & 55.3  \\
6 & 89.4 & 91.6 & 66.5  & 78.1 & 40.5 & 55.0 \\ \bottomrule
\end{tabular}
\vspace{-8pt}
\caption{Effect on mAP of \emph{number of masks} $T$~\eq{mask} in our attentional localization. Training on SfM-120k.}
\label{tab:thresholds}
\end{table}

\section{Conclusion}
\label{sec:conclusion}

We confirm that training and evaluation sets for instance-level image retrieval really should not have class overlap. Our new $\cR$GLDv2-clean dataset makes fair comparisons possible with previous clean datasets. The comparison between the two versions reveals that class overlap indeed brings inflated performance, although the relative difference in number of images is small. Importantly, the ranking of SOTA methods is different on the two training sets.

On the algorithmic front, D2R methods typically require an additional object detection training stage with location supervision, which is inherently inefficient. Our method \ours provides a single-stage training pipeline without the need for location supervision.
\ours improves the SOTA not only on established clean training sets but also on the newly released $\cR$GLDv2-clean.


\section*{{Acknowledgment}}
\vspace*{-4pt}
This work was supported by Institute of Information \& communications Technology Planning \& Evaluation (IITP) under the metaverse support program to nurture the best talents (IITP-2024-RS-2023-00254529) grant funded by the Korea government (MSIT).

 {\small
 \bibliographystyle{ieee_fullname}
 \bibliography{egbib}
}

\clearpage

\title{Supplementary material for \\ ``On Train-Test Class Overlap and Detection for Image Retrieval''}

\maketitle
\thispagestyle{empty}

 \setcounter{page}{1}


\appendix

\renewcommand{\theequation}{A\arabic{equation}}
\renewcommand{\thetable}{A\arabic{table}}
\renewcommand{\thefigure}{A\arabic{figure}}


\section{Implementation details}
\label{sec:Implementation}

In our experiments, we use a computational environment featuring 8 RTX 3090 GPUs with PyTorch\cite{Paszke01}. We perform transfer learning from models pre-trained on ImageNet~\cite{Russakovsky01}. To ensure a fair comparison with previous studies \cite{yang2021dolg, SCH01, tokens}, we configure the learning environment as closely as possible. Specifically, we use ResNet101~\cite{Zhang01} as the backbone with final feature dimension $d = 2048$.

We use ArcFace~\cite{Deng01} loss function for training, with margin parameter 0.3. For optimization, we use stochastic gradient descent with momentum 0.9, weight decay 0.00001, initial learning rate 0.001, a warm-up phase~\cite{He2018BagOT} of three epochs and cosine annealing. We train SfM-120k for 100 epochs and $\cR$GLDv2-clean for 50 epochs. Previous work has shown the effectiveness of preserving the original image resolution during the training of image retrieval models~\cite{Hao2016WhatIT, Gordo01}. We adopt this principle following~\cite{Yokoo01, SCH01, dtop}, where each training batch consists of images with similar aspect ratios instead of a single fixed size. The batch size is 128. Following DIR~\cite{Gordo01} and DELF~\cite{delf}, we carry out classification-based training of the backbone only and subsequently fine-tune the model. During fine-tuning, we train \ours while the backbone is frozen, as shown in \autoref{fig:A5_FT}.

For evaluation, we use multi-resolution representation~\cite{Gordo01} on both query and database images, applying $\ell_2$-normalization and whitening~\cite{Radenovic01} on the final features.

\begin{table}[h]
\centering
\scriptsize
\setlength{\tabcolsep}{5.5pt}
\begin{tabular}{lcc} \toprule
\Th{Network} & \Th{\#Params (M)} & \Th{\#GFLOPs} \\ \midrule
R101 & 42.50 & 7.86 \\
Yokoo~\etal~\cite{Yokoo01} & 43.91 & 7.86  \\
SOLAR~\cite{Ng01} & 53.36 & 8.57 \\
DOLG~\cite{yang2021dolg} & 47.07 & 8.07 \\
Token~\cite{tokens} & 54.43 & 8.05 \\ \midrule \rowcolor{LightCyan}
CiDeR (Ours) &  46.12 & 7.94 \\
\bottomrule
\end{tabular}
\vspace{-6pt}
\caption{\emph{Model complexity}: Parameters (\Th{\#Params}) and computational complexity (\Th{\#GFLOPs}) of different models providing official code. Single forward pass, given an input image of size 224 $\times$ 224.}
\label{tab:param}
\end{table}

\begin{table}[h]
\centering
\scriptsize
\setlength{\tabcolsep}{3.0pt}
\begin{tabular}{ccccccc} \toprule
\Th{R101} & \Th{BE} & \Th{SC} & \Th{AL} &\Th{ Pooling} & \Th{\#Params (M)} & \Th{\#GFLOPs} \\ \midrule
\ch &     &      &     &  & 42.50 & 7.86 \\
\ch & \ch &      &     &  & 43.58 & 7.91 \\
\ch & \ch & \ch  &     &  & 43.88 & 7.93 \\
\ch & \ch & \ch  & \ch &  & 44.02 & 7.94 \\
\ch & \ch & \ch  & \ch & \ch & 46.12 & 7.94 \\
\bottomrule
\end{tabular}
\vspace{-3pt}
\caption{\emph{Model complexity}: Parameters (\Th{\#Params}) and computational complexity (\Th{\#GFLOPs}) for different components of \ours. BE: backbone enhancement; SC: selective context; AL: attentional localization; Pooling: spatial pooling (GeM) + FC.}
\label{tab:param2}
\end{table}

\section{Model complexity}
\label{sec:Benchmark}

\autoref{tab:param} compares the model complexity\footnote{\url{https://github.com/sovrasov/flops-counter.pytorch}} of \ours with other models. In this table, R101 is the baseline for all related studies, all of which use the feature maps of its last layer. We observe that our model has the least complexity after Yokoo~\etal~\cite{Yokoo01}, which only uses GeM + FC. \autoref{tab:param2} shows model complexity for each of the components of \ours, as defined in \autoref{sec:impl}.

\begin{figure*}
\centering
\scriptsize
\setlength{\tabcolsep}{4pt}
\begin{tabular}{cc}
\fig[.5]{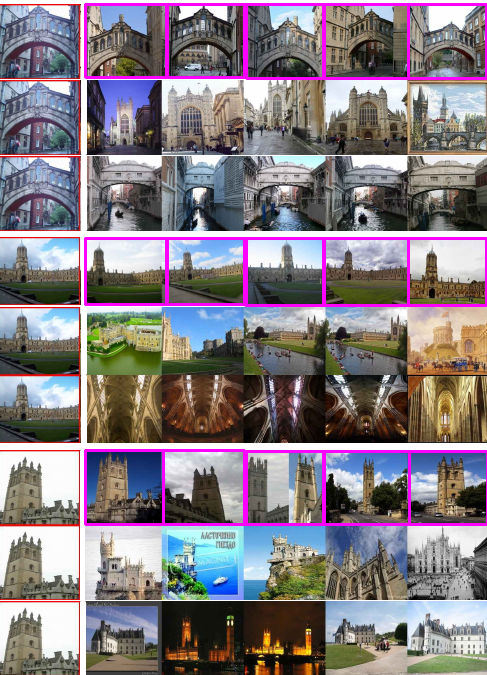} & \fig[.5]{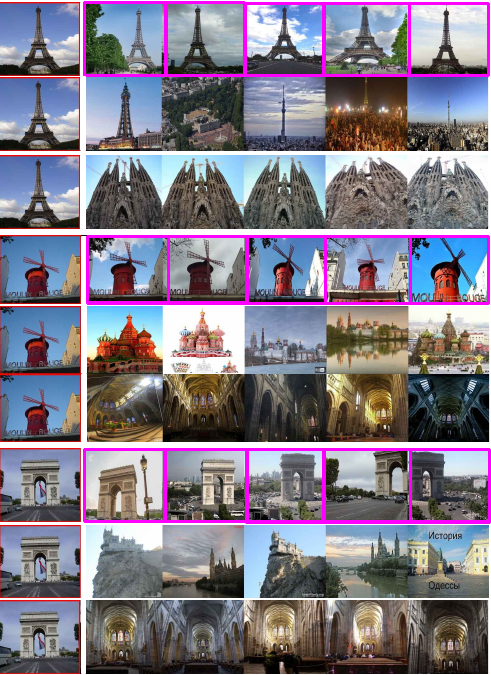} \\[3pt]
Evaluation: $\cR$Oxford & Evaluation: $\cR$Paris \\
\end{tabular}
\vspace{-6pt}
\caption{\emph{Confirming overlapping landmark categories} between training sets and evaluation sets ($\cR$Oxford, $\cR$Paris). Red box: query image. The query image from the evaluation set in each box/row is followed by top-5 most similar images from the training set (for each query, top down: GLDv2-clean, NC-clean, SfM-120k). Pink box: training image landmark identical with query (evaluation) image landmark.
}
\label{fig:fig2_sup}
\end{figure*}

\begin{table*}
\centering
\scriptsize
\begin{tabular}{cccll} \toprule
\Th{\#} & \Th{GID} & \Th{\# Images} & \Th{GLDv2 Landmark Name} & \Th{Oxford/Paris Landmark Name} \\
\midrule
 1 &   6190 & 98 & \href{http://commons.wikimedia.org/wiki/Category:Radcliffe_Camera}{Radcliffe Camera} & Oxford \\
 2 &  19172 & 32 & \href{http://commons.wikimedia.org/wiki/Category:All_Souls_College,_Oxford}{All Souls College, Oxford} & All Souls Oxford\\
 3 &  37135 & 18  & \href{http://commons.wikimedia.org/wiki/Category:Oxford_University_Museum_of_Natural_History}{Oxford University Museum of Natural History} & Pitt Rivers Oxford \\
 4 &  42489 & 55 & \href{http://commons.wikimedia.org/wiki/Category:Pont_au_Double}{Pont au Double} & Jesus Oxford\\
5 & 147275 & 18 & \href{http://commons.wikimedia.org/wiki/Category:Magdalen_Tower}{Magdalen Tower} & Ashmolean Oxford \\
6 & 152496 & 71 & \href{http://commons.wikimedia.org/wiki/Category:Christ_Church,_Oxford}{Christ Church, Oxford} & Christ Church Oxford\\
7 & 167275 & 55 & \href{http://commons.wikimedia.org/wiki/Category:Bridge_of_Sighs_(Oxford)}{Bridge of Sighs (Oxford)} & Magdalen Oxford \\
8 & 181291 & 60 & \href{http://commons.wikimedia.org/wiki/Category:Petit-Pont}{Petit-Pont} & Notre Dame Paris \\
9 & 192090 & 23 & \href{http://commons.wikimedia.org/wiki/Category:Christ_Church_Great_Quadrangle}{Christ Church Great Quadrangle} & Paris  \\
10 &  28949 & 91 & \href{http://commons.wikimedia.org/wiki/Category:Moulin_Rouge}{Moulin Rouge} & Moulin Rouge Paris \\
11 &  44923 & 41 & \href{http://commons.wikimedia.org/wiki/Category:Jardin_de_l'Intendant}{Jardin de l'Intendant} & Hotel des Invalides Paris \\
12 &  47378 & 731  & \href{http://commons.wikimedia.org/wiki/Category:Eiffel_Tower}{Eiffel Tower} & Eiffel Tower Paris \\ 
13 &  69195 & 34 & \href{http://commons.wikimedia.org/wiki/Category:Place_Charles-de-Gaulle_(Paris)}{Place Charles-de-Gaulle (Paris)} & Arc de Triomphe Paris \\ 
14 & 167104 & 23 & \href{http://commons.wikimedia.org/wiki/Category:H%C3%B4tel_des_Invalides}{Hôtel des Invalides} & Hotel des Invalides Paris \\
15	& 145268 & 72 & \href{http://commons.wikimedia.org/wiki/Category:Louvre_Pyramid}{Louvre Pyramid} & Louvre Paris\\
16	& 146388 & 80 & \href{http://commons.wikimedia.org/wiki/Category:Basilique_du_Sacr%C3%A9-C%C5%93ur_de_Montmartre}{Basilique du Sacré-Cœur de Montmartre} & Arc de Triomphe Paris  \\
17 & 138332 & 30 & \href{http://commons.wikimedia.org/wiki/Category:Parvis_Notre-Dame_-_place_Jean-Paul-II_(Paris)}{Parvis Notre-Dame - place Jean-Paul-II (Paris)} & Notre Dame Paris  \\
18 & 144472 & 33 & \href{http://commons.wikimedia.org/wiki/Category:Esplanade_des_Invalides}{Esplanade des Invalides} & Paris \\
\bottomrule
\end{tabular}
\vspace{-6pt}
\caption{Details of GIDs removed from GLDv2-clean dataset.}
\label{tab:removing_gid}
\end{table*}


\section{More on revisited \vs original GLDv2-clean}

\paragraph{Details}

To identify overlapping landmarks, we use GLAM~\cite{SCH01} to extract image features from the training and evaluation sets. Extracted features from the training sets are indexed using the Approximate Nearest Neighbor (ANN)\footnote{\url{https://github.com/kakao/n2}} search method. For verification, we use SIFT~\cite{Lowe01} local descriptors. We find tentative correspondences between local descriptors by a kd-tree and we verify by obtaining inlier correspondences using RANSAC.

In addition to \autoref{fig:fig2} in~\autoref{sec:dataset}, \autoref{fig:fig2_sup} shows overlapping landmark categories between the training set (GLDv2-clean, NC-clean, SfM-120k) and the evaluation set ($\cR$Oxford, $\cR$Paris). Clearly, only GLDv2-clean has overlapping categories with the evaluation set.

\autoref{tab:removing_gid} shows the details of the 18 GIDs that are removed from GLDv2-clean due to overlap with the evaluation sets. The new, revisited $\cR$GLDv2-clean dataset is what remains after this removal.

\begin{table}[h]
\centering
\tiny
\setlength{\tabcolsep}{2.5pt}
\begin{tabular}{l*{10}{c}} \toprule
\mr{2}{\Th{Method}}& \mr{2}{\Th{trainset}} & \mr{2}{\Th{OC}} & \mr{2}{\Th{\oxf5k}} & \mr{2}{\Th{\paris6k}} & \mc{2}{\Th{Medium}} & \mc{2}{\Th{Hard}} & \mr{2}{\Th{Mean}} &\\ \cmidrule(l){6-9}
 & & & & & \rox & \rpa & \rox & \rpa \\ \midrule
\mr{2}{SOLAR~\cite{tokens}}                & \mr{2}{GLDv2-clean}      & Y & 82.1 & 95.2 & 72.5 & 88.8 & 47.3 & 75.8 & 77.0 \\
                                           &                          & N & 81.6 & 95.9 & 66.1 & 84.8 & 44.3 & 70.6 & 73.9 \\ 
\mr{2}{SOLAR~\cite{Ng01}$^\dagger$}        & \mr{2}{$\cR$GLDv2-clean} & Y & 77.7 & 87.6 & 65.4 & 78.4 & 36.0 & 62.2 & 67.9 \\
                                           &                          & N & 80.1 & 92.0 & 66.3 & 81.6 & 42.6 & 68.7 & 71.9 \\ \midrule
\mr{2}{GLAM~\cite{SCH01}}                  & \mr{2}{GLDv2-clean}      & Y & 81.6 & 93.9 & 73.6 & 88.6 & 53.6 & 77.4 & 78.1 \\
                                           &                          & N & 76.8 & 94.2 & 62.9 & 83.8 & 42.0 & 69.5 & 71.5 \\ 
\mr{2}{GLAM~\cite{SCH01}$^\ddagger$}       & \mr{2}{$\cR$GLDv2-clean} & Y & 76.4 & 89.4 & 69.3 & 85.2 & 48.9 & 74.2 & 73.9 \\
                                           &                          & N & 75.6 & 93.3 & 61.7 & 84.0 & 43.1 & 68.1 & 71.0 \\ \midrule
\mr{2}{DOLG~\cite{dtop}}                   & \mr{2}{GLDv2-clean}      & Y & 81.5 & 94.3 & 72.8 & 87.0 & 48.2 & 76.0 & 76.6 \\
                                           &                          & N & 75.7 & 93.1 & 62.7 & 82.1 & 42.0 & 64.4 & 70.0 \\ 
\mr{2}{DOLG~\cite{yang2021dolg}$^\dagger$} & \mr{2}{$\cR$GLDv2-clean} & Y & 76.1 & 88.6 & 66.1 & 79.7 & 41.5 & 64.1 & 69.4 \\
                                           &                          & N & 74.6 & 91.9 & 61.1 & 82.0 & 37.1 & 65.0 & 68.6 \\
\bottomrule
\end{tabular}
\vspace{-5pt}
\caption{mAP comparison of the original GLDv2-clean training set with our revisited version $\cR$GLDv2-clean separately for \emph{overlapping classes (OC) \vs non-overlapping} for a number of SOTA methods. For GLDv2-clean, we evaluate pre-trained models. For $\cR$GLDv2-clean we reproduce training with ResNet101 backbone, ArcFace loss and same sampling, settings and hyperparameters. $\dagger/\ddagger$: official/our code. }
\vspace{-10pt}
\label{tab:overlap}
\end{table}

\paragraph{Classes with/without overlap}

\autoref{tab:overlap} elaborates on the results of \autoref{tab:sota2_diff} by comparing the original GLDv2-clean training set with our revisited version $\cR$GLDv2-clean separately for overlapping \vs non-overlapping classes. That is, classes of the evaluation set that overlap or not with the original training set. As expected, mAP is much higher for overlapping than non-overlapping classes on GLDv2-clean. On $\cR$GLDv2-clean, differences are smaller or even non-overlapping are higher.

\begin{figure*}
\begin{center}
	\fig[1.0]{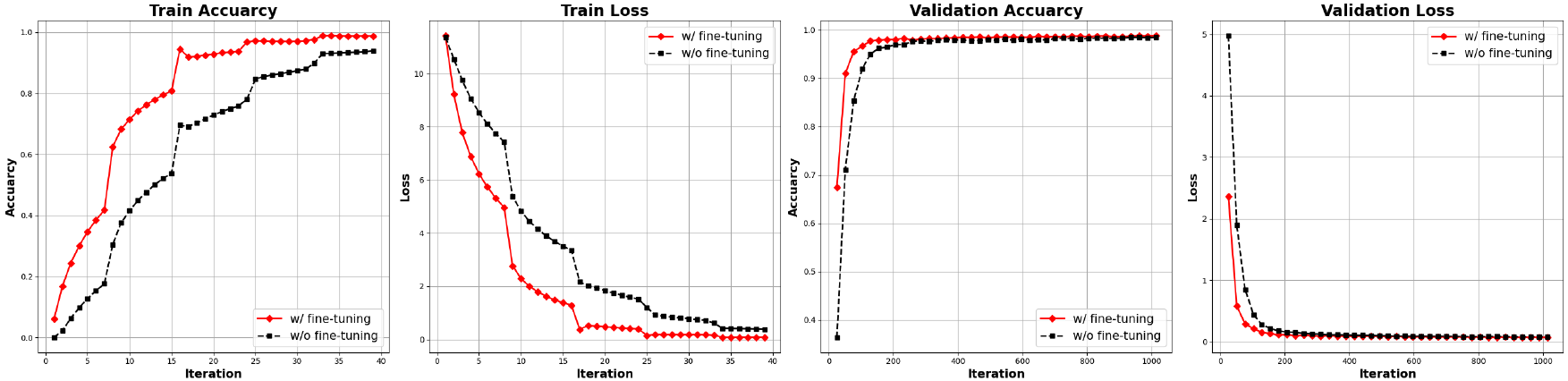}
\end{center}
\vspace{-12pt}
\caption{Comparison of the accuracy and loss for training and validation \emph{with (red) \vs without (black) fine-tuning}.}
\label{fig:ft}
\end{figure*}

\begin{figure}
\centering
\fig[.9]{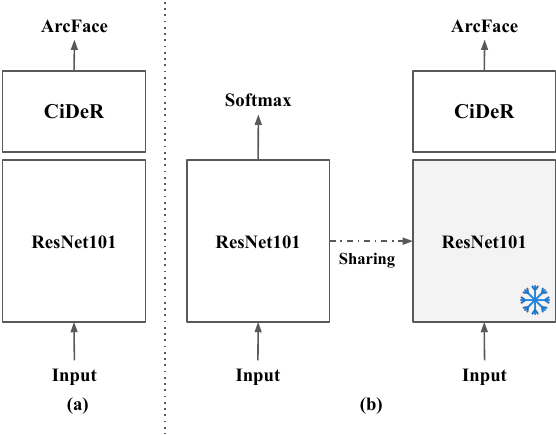}
\caption{\emph{Fine-tuning process}. (a) No fine-tuning. (b) Our fine-tuning. \includegraphics[height=0.3cm]{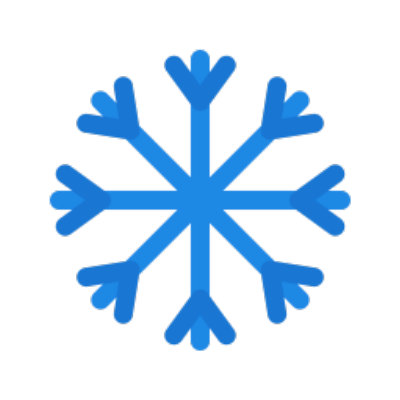}: frozen.}
\vspace{-12pt}
\label{fig:A5_FT}
\end{figure}

\section{More ablations}
\label{sec:mAblation}

\paragraph{Fine-tuning}

We employ transfer learning from models pre-trained on ImageNet~\cite{Russakovsky01}. Following DIR~\cite{Gordo01} and DELF~\cite{delf}, we first perform classification-based training of the backbone only on the landmark training set and then fine-tuning the model on the same training set, training \ours while the backbone is kept frozen. \autoref{fig:A5_FT} visualizes this process, while \autoref{fig:ft} shows the training and validation loss and accuracy, with and without the fine-tuning process. These plots confirm that fine-tuning results in lower loss and higher accuracy on both training and validation sets. This is corroborated by improved performance results ({\ours+FT}) in \autoref{tab:1m}. Compared to the results without the fine-tuning, we obtain gains of 2.7\% and 3.1\% on \oxf5k and \paris6k Base, 8.9\% and 5.1\% on \rox and \rpa Medium, and 16.5\% and 11.4\% on \rox and \rpa Hard.

\begin{table}
\centering
\scriptsize
\setlength{\tabcolsep}{5.0pt}
\begin{tabular}{l*{7}{c}} \toprule
\mr{2}{\Th{Method}} & \mr{2}{\Th{Oxf5k}} & \mr{2}{\Th{Par6k}} & \mc{2}{\Th{Medium}} & \mc{2}{\Th{Hard}} \\ \cmidrule(l){4-7}
 & & &  \rox & \rpa & \rox & \rpa \\ \midrule
ECNet~\cite{wang01} &  88.2     & {91.5} & {66.8}      & {78.3}       & {42.0}       & {55.4}       \\
NLNet~\cite{Wang02}           & 89.4      & {91.8} &  {66.5}     & {77.6}      & {39.1}       & {53.7}  \\
Gather-Excite~\cite{GatherExcite} &  89.4     & {90.5} &  {66.7}     & {77.1}       & {41.2}    & {53.8}  \\
\rowcolor{LightCyan}
SENet~\cite{Hu01}          & \tb{89.9} & \tb{92.0} & {\tb{67.3}} &  {\tb{79.4}} & {\tb{42.4}} & {\tb{57.5}}  \\
\bottomrule
\end{tabular}
\vspace{-5pt}
\caption{mAP comparison of different \emph{backbone enhancement} (BE) options. Training on SfM-120k.}
\label{tab:senet}
\end{table}

\paragraph{Backbone enhancement (BE)}

We apply four methods in a plug-and-play fashion~\cite{wang01, Wang02, GatherExcite, Hu01}. As shown in \autoref{tab:senet}, SENet~\cite{Hu01} performs best. We select it for backbone enhancement in the remaining experiments.

\begin{table}
\centering
\scriptsize
\setlength{\tabcolsep}{6.0pt}
\begin{tabular}{l*{7}{c}} \toprule
\mr{2}{\Th{Method}} & \mr{2}{\Th{Oxf5k}} & \mr{2}{\Th{Par6k}} & \mc{2}{\Th{Medium}} & \mc{2}{\Th{Hard}} \\ \cmidrule(l){4-7}
 & & &  \rox & \rpa & \rox & \rpa \\ \midrule
ASPP~\cite{Chen2017RethinkingAC}  & \tb{90.3}   & 92.2    &  \tb{67.9} & 78.2 & 41.6 & 55.8\\
SKNet~\cite{sknet} & 89.3  & \tb{92.4} & {67.4} & 78.4 & 42.3 & 55.5 \\
\rowcolor{LightCyan}
SKNet$^\dagger$ & {89.9} & {92.0} & {67.3} &  {\tb{79.4}}& {\tb{42.4}} & {\tb{57.5}}\\ \bottomrule
\end{tabular}
\vspace{-5pt}
\caption{mAP comparison of different \emph{selective context} (SC) options. Training on SfM-120k. SKNet$^\dagger$: our modification of SKNet~\cite{sknet}.}
\label{tab:scm}
\end{table}

\paragraph{Selective context (SC)}

Here we compare ASPP~\cite{Chen2017RethinkingAC}, SKNet~\cite{sknet} and our modification, SKNet$^\dagger$. The modification is that instead of a simple \emph{element-wise sum} to initially fuse multiple context information, we introduce a \emph{learnable parameter}~\eq{fuse} to fuse feature maps based on importance. As shown in \autoref{tab:scm}, our modification SKNet$^\dagger$ performs best, confirming that this approach better embeds context information.

\begin{table}[t]
\centering
\scriptsize
\setlength{\tabcolsep}{5.5pt}
\begin{tabular}{l*{8}{c}} \toprule
\mr{2}{SC} & \mr{2}{AL} & \mr{2}{\Th{Oxf5k}} & \mr{2}{\Th{Par6k}} & \mc{2}{\Th{Medium}} & \mc{2}{\Th{Hard}} \\ \cmidrule(l){5-8}
    &      &           &            & \rox       & \rpa      & \rox      & \rpa      \\ \midrule
    &      &   87.1    & 90.6       & 63.9       & 77.3      & 36.7      & 53.9      \\
\ch &      &   87.6    & 90.8       & 64.7       & 77.8      & 37.9      & 54.8      \\
    & \ch  &   89.7    & 92.0       & 66.8       & \tb{79.4} & 41.8      & \tb{57.5} \\
\ch & \ch  & \tb{89.9} & \tb{92.0}  & \tb{67.3}  & \tb{79.4} & \tb{42.4} & \tb{57.5} \\
\bottomrule
\end{tabular}
\vspace{-8pt}
\caption{mAP comparison of \emph{learnable-fusion} (\ch) \vs \emph{sum}. Training on SfM-120k. SC: selective context; AL: attentional localization.}
\label{tab:learnablefusing}
\end{table}

\paragraph{Sum (baseline) \vs learnable fusion}

We introduce learnable parameters~\eq{fuse} to fuse multiple feature maps for SC and AL. \autoref{tab:learnablefusing} compares this \emph{learnable fusion} with simple \emph{sum} for both SC and AL. We evaluate four different combinations, using learnable fusion and sum for SC and AL. The results indicate that learnable fusion improves performance wherever it is applied.


\begin{table}[t]
\centering
\scriptsize
\setlength{\tabcolsep}{3.0pt}
\begin{tabular}{l*{7}{c}} \toprule
\mr{2}{\Th{Backbone}} & \mr{2}{\Th{Oxf5k}} & \mr{2}{\Th{Par6k}} & \mc{2}{\Th{Medium}} & \mc{2}{\Th{Hard}} \\ \cmidrule(l){4-7}
 & & &  \rox & \rpa & \rox & \rpa \\ \midrule
Attention-based pooling           & 89.8      & \tb{92.3} & 67.2      &  \tb{79.4}    & 41.8      & 56.5  \\
\rowcolor{LightCyan}
Mask-based pooling (Ours) & \tb{89.9} & {92.0}    & \tb{67.3}&  \tb{79.4}& \tb{42.4} & \tb{57.6}\\ \bottomrule
\end{tabular}
\vspace{-5pt}
\caption{mAP comparison of pre-trained model with \emph{attention-based \vs mask-based pooling}. Training on SfM-120k.}
\label{tab:pooling}
\end{table}

\paragraph{Attention-based \vs mask-based pooling}

Because of the binary masks~\eq{mask}, the pooling operation of our attentional localization (AL) can be called \emph{mask-based pooling}. Here we derive a simpler baseline and connect it with attention in transformers. Given the feature tensor $\vF \in \real^{w \times h \times d}$, we flatten the spatial dimensions to obtain the \emph{keys} $K \in \real^{p \times d}$, where $p = w \times h$ is the number of patches. The weights of the $1 \times 1$ convolution $f^\ell$ can be represented by \emph{query} $Q \in \real^{1 \times d}$, which plays the role of a learnable CLS token. Then, replacing the nonlinearity $\eta(\zeta(\cdot))$ by softmax, the spatial attention map~\eq{attn} becomes
\begin{equation}
	A = \softmax(Q K\tran) \in \real^{1 \times p}.
\label{eq:attn2}
\end{equation}
Then, by omitting the masking operation and using the attention map $A$ to weight the \emph{values} $V = K \in \real^{p \times d}$, \eq{alm} simplifies to
\begin{equation}
	\vF^\ell = A\tran \odot V \in \real^{p \times d},
\label{eq:alm2}
\end{equation}
Finally, we apply spatial pooling $f^p$, like GAP or GeM. For example, in the case of GAP, the final pooled representation becomes
\begin{equation}
	f^p(\vF^\ell) = A V \in \real^{1 \times d},
\label{eq:att-pool}
\end{equation}
which is the same as a simplified cross-attention operation between the features $\vF$ and a learnable CLS token, without projections. By using GeM pooling, we refer to this baseline as \emph{attention-based pooling}. Variants of this approach have been used, mostly for classification~\cite{lee2019set, zhai2022scaling, pmlr-v139-radford21a, touvron2021augmenting, psomas2023simpool}. As shown in \autoref{tab:pooling}, our mask-based pooling is on par or performs better than the attention-based pooling baseline, especially on the hard protocol.

\begin{table}[t]
\centering
\scriptsize
\setlength{\tabcolsep}{8.0pt}
\begin{tabular}{*{7}{c}} \toprule
\mr{2}{\Th{Dim}} & \mr{2}{\Th{Oxf5k}} & \mr{2}{\Th{Par6k}} & \mc{2}{\Th{Medium}} & \mc{2}{\Th{Hard}} \\ \cmidrule(l){4-7}
 & & &  \rox & \rpa & \rox & \rpa \\ \midrule
4096 & 87.8 & 90.2 & 64.8 &  76.8 & 38.1 & 52.8 \\
3097 & 89.8 & 90.5 & \tb{67.4} &  76.9 & \tb{42.5} & 53.0 \\
\rowcolor{LightCyan}
2048 & \tb{89.9} & \tb{92.0} & {67.3} &  {\tb{79.4}}& {{42.4}} & {\tb{57.5}}\\
1024 & 88.9  & 91.2  & 65.7 &  76.7 & 40.1 & 52.0 \\
512  & 85.3  & 89.2  & 61.9 &  74.4 & 36.5 & 48.9 \\
\bottomrule
\end{tabular}
\vspace{-5pt}
\caption{mAP comparison of different \emph{feature dimensions} $d$ in our model. Training on SfM-120k.}
\label{tab:dimension}
\end{table}

\paragraph{Feature dimension}

After applying spatial pooling like GeM, we apply an FC layer to generate the final features. The feature dimension $d$ is a hyperparameter. \autoref{tab:dimension} shows the performance for different dimensions $d$. Interestingly, a feature dimension of 2,048 works best, with larger dimensions not necessarily offering any more performance improvement.

\begin{table}
\centering
\scriptsize
\setlength{\tabcolsep}{4.0pt}
\begin{tabular}{l*{7}{c}} \toprule
\mr{2}{\Th{Query}} & \mr{2}{\Th{Database}} & \mr{2}{\Th{Oxf5k}} & \mr{2}{\Th{Par6k}} & \mc{2}{\Th{$\cR$Medium}} & \mc{2}{\Th{$\cR$Hard}} \\ \cmidrule(l){5-8}
 & & & & \rox & \rpa & \rox & \rpa \\ \midrule
Single & Single &  90.5 & 91.5 & 67.0 & 77.4 & 40.3 & 55.0 \\
Multi & Single &  \tb{92.6} & \tb{92.9} & \tb{68.4} & {79.2} & 41.2 & 56.5 \\
Single & Multi & 87.1 & 90.4 & 64.8 & 77.5 & 39.1 & 55.9 \\
\rowcolor{LightCyan}
Multi & Multi & {89.9} & {92.0} & {67.3} &  {\tb{79.4}}& {\tb{42.4}} & {\tb{57.5}}\\  \bottomrule
\end{tabular}
\vspace{-6pt}
\caption{mAP comparison using \emph{multiresolution} representation (Multi) or not (Single) on query or database images. Training on SfM-120k.}
\label{tab:QE}
\end{table}

\paragraph{Multi-resolution}

At inference, we use a multi-resolution representation at image scales (0.4, 0.5, 0.7, 1.0, 1.4) for both the query and the database images. Features are extracted at each scale and then averaged to form the final representation. \autoref{tab:QE} provides a comparative analysis, with and without the multi-resolution representation for query and database images. We find that applying multi-resolution to both query and database images works best for \roxf and \rpar~\cite{RITAC18}.


\begin{table}[t]
\centering
\scriptsize
\setlength{\tabcolsep}{6.0pt}
\begin{tabular}{l*{7}{c}} \toprule
\mr{2}{\Th{Backbone}} & \mr{2}{\Th{Oxf5k}} & \mr{2}{\Th{Par6k}} & \mc{2}{\Th{Medium}} & \mc{2}{\Th{Hard}} \\ \cmidrule(l){4-7}
 & & &  \rox & \rpa & \rox & \rpa \\ \midrule
Facebook           & 88.2      & \tb{92.7} & 65.3      &  78.8    & 38.6      & 56.5  \\
\rowcolor{LightCyan}
TorchVision & \tb{89.9} & {92.0}    & \tb{67.3}&  \tb{79.4}& \tb{42.4} & \tb{57.5}\\ \bottomrule
\end{tabular}
\vspace{-5pt}
\caption{mAP comparison of pre-trained model from \emph{TorchVision \vs Facebook}. Training on SfM-120k.}
\label{tab:backbone}
\end{table}

\paragraph{ImageNet pre-trained models}

Different research teams have released models pre-trained on ImageNet~\cite{Russakovsky01} for major image classification tasks. It is common to use a pre-trained ResNet101 model from TorchVision~\cite{torchvision}. Recent works~\cite{yang2021dolg, cvnet} have also used pre-trained models released by Facebook\footnote{\url{https://github.com/facebookresearch/pycls}}. As shown in \autoref{tab:backbone}, we find that the TorchVision model works best.

\begin{table}[t]
\centering
\scriptsize
\setlength{\tabcolsep}{5.6pt}
\begin{tabular}{*{7}{c}} \toprule
\mr{2}{\Th{Warm-up}} & \mr{2}{\Th{Oxf5k}} & \mr{2}{\Th{Par6k}} & \mc{2}{\Th{Medium}} & \mc{2}{\Th{Hard}} \\ \cmidrule(l){4-7}
 & & &  \rox & \rpa & \rox & \rpa \\ \midrule
    & \tb{89.9}  & \tb{92.0} & 66.7      &  78.9      & 41.5      & 56.7  \\
\rowcolor{LightCyan}
\ch & \tb{89.9}  & \tb{92.0} & \tb{67.3} &  \tb{79.4} & \tb{42.4} & \tb{57.5}\\ \bottomrule
\end{tabular}
\vspace{-5pt}
\caption{mAP effect of \emph{warm-up} in our model training. Training on SfM-120k.}
\label{tab:Warmup}
\end{table}

\paragraph{Warm-Up}

To enhance model performance, we employ a warm-up phase~\cite{He2018BagOT} during training, consisting of three epochs. \autoref{tab:Warmup} shows that the warm-up phase improves the performance.

\begin{table}[t]
\centering
\scriptsize
\setlength{\tabcolsep}{5.3pt}
\begin{tabular}{*{7}{c}} \toprule
\mr{2}{\Th{whitening}} & \mr{2}{\Th{Oxf5k}} & \mr{2}{\Th{Par6k}} & \mc{2}{\Th{Medium}} & \mc{2}{\Th{Hard}} \\ \cmidrule(l){4-7}
 & & &  \rox & \rpa & \rox & \rpa \\ \midrule
    & 85.8      & 90.7      & 60.5      &  77.0    & 31.7      & 54.2  \\
\rowcolor{LightCyan}
\ch & \tb{89.9} & \tb{92.0}     & \tb{67.3} & \tb{79.4}& \tb{42.4} & \tb{57.5}\\ \bottomrule
\end{tabular}
\vspace{-5pt}
\caption{mAP effect of \emph{whitening} in our model. Training on SfM-120k.}
\label{tab:whitening}
\end{table}

\paragraph{Whitening}

We utilize the supervised whitening method pioneered by Radenović~\etal~\cite{Radenovic01}, which is common in related work to improve retrieval performance. \autoref{tab:whitening} shows the performance gain obtained by the application of whitening.

\begin{table}
\centering
\scriptsize
\setlength{\tabcolsep}{4.2pt}
\begin{tabular}{l*{6}{c}} \toprule
\mr{2}{\Th{Method}} & \mr{2}{\Th{Oxf5k}} & \mr{2}{\Th{Par6k}} & \mc{2}{\Th{$\cR$Medium}} & \mc{2}{\Th{$\cR$Hard}} \\ \cmidrule(l){4-7}
 & & & \rox & \rpa & \rox & \rpa \\ \midrule
Fixed-size (224 × 224) & 69.0 & 86.0 & 42.2 & 69.5 & 15.8 & 45.0\\
\rowcolor{LightCyan}
Group-size (Ours) & \tb{89.9}  & \tb{92.0} & \tb{67.3} &  \tb{79.4} & \tb{42.4} & \tb{57.5} \\
\bottomrule
\end{tabular}
\vspace{-6pt}
\caption{mAP comparison between \emph{fixed-size} ($224 \times 224$) \vs \emph{group-size sampling}. Training on SfM-120k.}
\label{tab:groupsize}
\end{table}

\paragraph{Fixed-size \vs group-size sampling}

Several previous studies suggest organizing training batches based on image size for efficient learning.
Methods such as DIR~\cite{Gordo01}, DELF~\cite{delf}, MobileViT~\cite{mehta2021mobilevit}, and Yokoo~\etal\cite{Yokoo01} opt for variable image sizes rather than adhering to a single, fixed dimension.
Our approach employs group-size sampling~\cite{Yokoo01, SCH01, dtop}, where we construct image batches with similar aspect ratios. \autoref{tab:groupsize} compares the results of fixed-size (224 $\times$ 224) and group-size sampling.
We find that using dynamic input sizes to preserve the aspect ratio significantly improves performance.



\end{document}